\documentclass[10pt,twocolumn,letterpaper]{article}

\usepackage[pagenumbers]{cvpr} 

\definecolor{cvprblue}{rgb}{0.21,0.49,0.74}
\usepackage[pagebackref,breaklinks,colorlinks,allcolors=cvprblue]{hyperref}

\usepackage{amsthm}
\usepackage{bbm}
\usepackage{siunitx}
\usepackage{threeparttable}
\usepackage{multirow}
\usepackage{makecell}
\usepackage{colortbl}
\usepackage{algorithm}
\usepackage{algorithmic}
\usepackage{listings}

\usepackage{amssymb}
\usepackage{xcolor}
\usepackage{subcaption}

\definecolor{ourcolor}{HTML}{3B86B9}
\colorlet{ourcolor}{ourcolor!10}

\def\paperID{{*****}} 
\def\confName{CVPR}
\def\confYear{2026}

\title{Stable Spike: Dual Consistency Optimization via Bitwise AND Operations for Spiking Neural Networks}

\author{
Yongqi Ding, Kunshan Yang, Linze Li, Yiyang Zhang, Mengmeng Jing, Lin Zuo\thanks{Corresponding author (linzuo@uestc.edu.cn).}\\
School of Information and Software Engineering\\
University of Electronic Science and Technology of China\\
}

\begin{document}
\maketitle
\begin{abstract}
Although the temporal spike dynamics of spiking neural networks (SNNs) enable low-power temporal pattern capture capabilities, they also incur inherent inconsistencies that severely compromise representation. In this paper, we perform dual consistency optimization via \textbf{Stable Spike} to mitigate this problem, thereby improving the recognition performance of SNNs. With the hardware-friendly ``AND"  bit operation, we efficiently decouple the stable spike skeleton from the multi-timestep spike maps, thereby capturing critical semantics while reducing inconsistencies from variable noise spikes. Enforcing the unstable spike maps to converge to the stable spike skeleton significantly improves the inherent consistency across timesteps. Furthermore, we inject amplitude-aware spike noise into the stable spike skeleton to diversify the representations while preserving consistent semantics. The SNN is encouraged to produce perturbation-consistent predictions, thereby contributing to generalization. Extensive experiments across multiple architectures and datasets validate the effectiveness and versatility of our method. In particular, our method significantly advances neuromorphic object recognition under ultra-low latency, improving accuracy by up to 8.33\%. This will help unlock the full power consumption and speed potential of SNNs.

\end{abstract}

\section{Introduction}
The rapid development of deep learning, especially the emergence of large models, makes the power consumption of AI models a non-negligible challenge. The human brain, on the other hand, enjoys a significant power consumption advantage over artificial neural networks (ANNs)~\cite{roy2019towards}, making neuromorphic computing a promising low-power, high-performance computing paradigm~\cite{MAASS19971659}. Spiking neural networks (SNNs) transmit sparse binary spikes over multiple timesteps, requiring only addition operations when deployed on neuromorphic chips, and consuming significantly less power compared to ANNs~\cite{yao2024spike,Spike_driven_Transformer,yang2024vision}. Similarly, neuromorphic data represents information as sparse binary events, and recognizing neuromorphic objects using SNNs can be inherently low-power and low-latency~\cite{yao2024spike,9543525,SSNN}.

However, although temporal spike dynamics give SNNs low-power spatio-temporal representations, they also introduce the risk of inconsistency. Differences in neuronal states and input currents across timesteps lead to excessive variability in spike maps and predictions, negatively affecting overall performance~\cite{MPS}. To address this problem, \cite{MPS} indirectly promotes spike consistency through membrane potential smoothing and maintains the overall stability  by performing logit distillation with adjacent timesteps. Although effective, this indirect strategy requires the modification of neuronal dynamics and struggles to be readily adopted as a versatile SNN enhancement solution, especially for deployment on neuromorphic chips where neuron models are often predetermined~\cite{yao2024spike,9900453,10.1093/nsr/nwae102}. Therefore, efficiently and versatilely promoting the predictive consistency and performance of SNNs remains an open challenge.

\begin{figure*}
  \centering
  \includegraphics[width=0.93\linewidth]{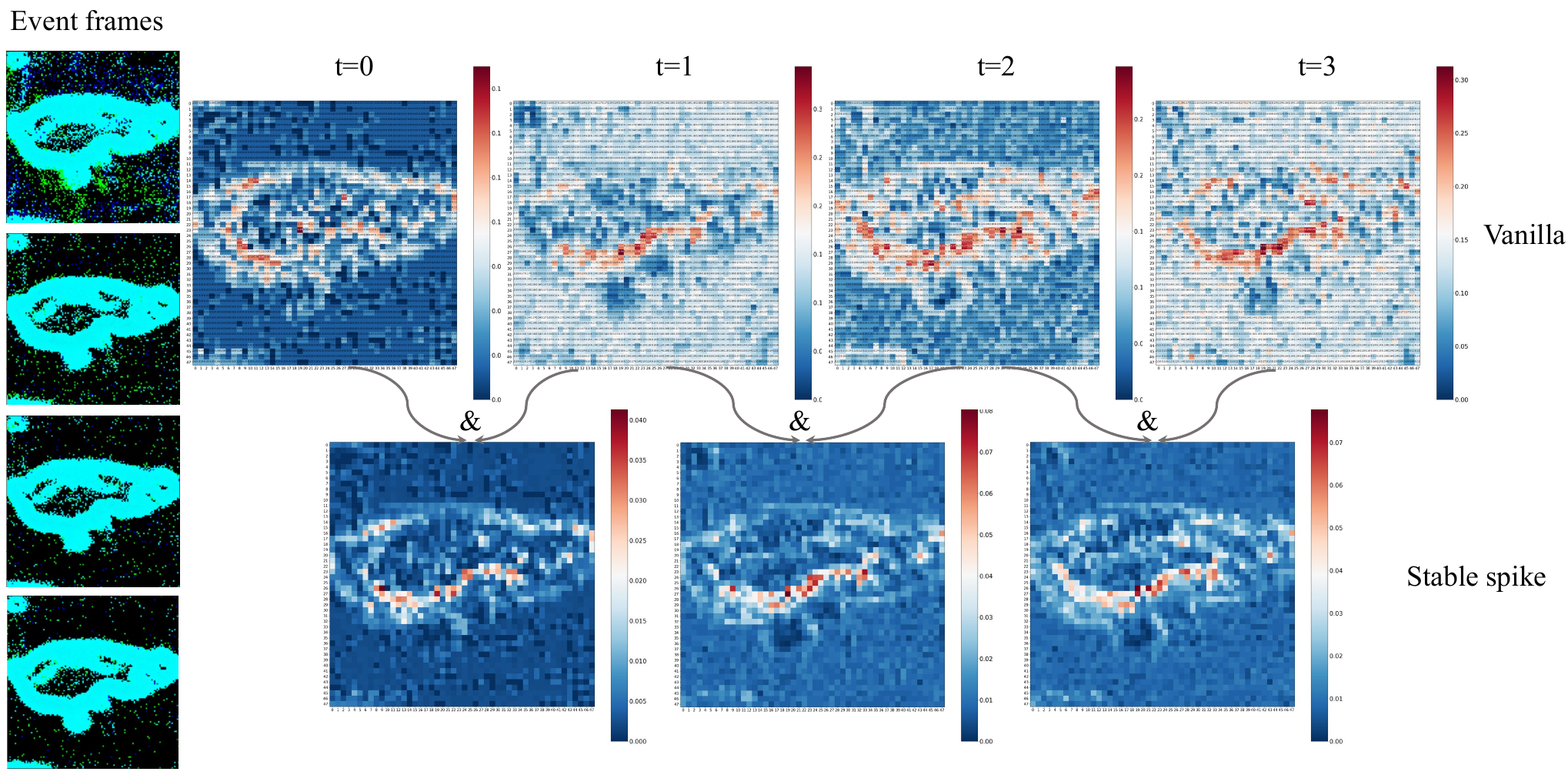}
  \vskip -0.06in
  \caption{Comparison of vanilla SNN spike maps and stable spike maps. Vanilla spike maps varied widely across timesteps, negatively affecting the overall representation; stable spike maps, decoupled by minimal \& operation, consistently represented the feature skeleon. The event frames are from the CIFAR10-DVS dataset and the visualization shows the spike maps of the first layer in VGG-9 after averaging over all channels. Additional visualizations in \textbf{Supplementary Material} show that the stable spike consistently extracts the feature skeleton.}
  \label{fig1}
\vskip -0.22in
\end{figure*}

In this paper, we propose the \textbf{Stable Spike}, which efficiently improves the consistency and performance of the SNN without any modification to the spiking neurons or architecture. Our method stems from a key insight: \textit{despite differences, a well-optimized SNN can capture object-critical features at different timesteps}. This is illustrated by the spike map visualization in Fig.~\ref{fig1}, where spike maps of different timesteps capture object-relevant features, while excessive redundant noise spikes unrelated to the object lead to variability. Therefore, we can decouple and extract the critical information, i.e. the stable spike skeleton, from the variational spike maps. To do this, we perform a hardware-friendly ``AND" bit operation~\cite{9923835,9731545} on the spike maps of adjacent timesteps to efficiently retrieve consistent and stable 1-value spikes while ignoring messy spikes. As shown in Fig.~\ref{fig1}, the stable spike maps reduce variability and outline a clear feature skeleton. We encourage the original spike maps to converge to the stable spike during training, which directly reduces the discrepancy between the multi-timestep spike maps and mitigates the interference of redundant, varying noise spikes.

While consistency across timesteps stabilizes the overall prediction, moderate diversity promotes generalization and convergence~\cite{Li_2021_ICCV,gao2022hyperbolic}. However, unlike ANNs that can directly benefit from random Gaussian noise~\cite{Li_2021_ICCV}, SNNs necessitate that (1) the noise to be discrete or else a mismatch in training-inference precision will occur, and (2) the noise amplitude to be appropriate or else it will interfere with normal training. To this end, we propose amplitude-aware spike noise, which possesses the discrete properties of binary spikes and generates noise depending on the amplitude of the spike firing rate. This enables the noise to adequately contribute to the generalization of high-amplitude elements while preventing low-amplitude elements from being overly perturbed. We inject amplitude-aware spike noise into the stable spike firing rate skeleton to preserve critical object semantics and push the perturbed predictions to approximate the original predictions. This enhances the perturbation consistency of SNNs to unknown variations, thereby improving generalizability.

Facilitated by dual consistency, our method efficiently boosts the performance of the SNN without changing the neurons or the model architecture. This allows our method to be plug-and-play for different spiking neurons and architectures, and to synergize with other methods to enhance performance (as shown in Table~\ref{tab:generalization}) with excellent versatility. In particular, our method significantly enhances performance on neuromorphic datasets, boosting the accuracy of DVS-Gesture by up to 8.33\% under ultra-low latency at two timesteps. Our contributions are summarized below: 
\begin{itemize}
    \item We propose to decouple the stable spike feature skeleton across timesteps from the SNN by the efficient ``AND" operation, and provide consistency guidance to the unstable spike maps.
    \item We propose to inject amplitude-aware spike noise into the stable spike firing rate to increase feature diversity and improve generalization by enhancing the perturbation prediction consistency of the SNN.
    \item Extensive experiments across multiple architectures and datasets demonstrate the effectiveness and versatility of our method, especially at enabling neuromorphic object recognition with ultra-low latency.
\end{itemize}

\vspace{-0.15in}
\section{Related Work}
\vspace{-0.05in}
\textbf{Spiking Neural Network.} SNNs mimic the information transfer mechanism of the biological nervous system, where binary spike signals are transmitted between spiking neurons over multiple timesteps~\cite{roy2019towards,Spike_driven_Transformer,9543525}. Additionally, spiking neuron dynamics can capture implicit temporal dependencies, enabling SNNs to exhibit superior spatio-temporal properties~\cite{ponghiran2022spiking}. This allows SNNs to be efficiently deployed on neuromorphic chips, providing performance comparable to that of ANNs while consuming substantially less power~\cite{yao2024spike,yang2024vision}. Previous studies have focused on improving the neuron dynamics~\cite{PSN,CLIF}, model architectures~\cite{Spike_driven_Transformer,QKFormer,wang2025spiking}, and training algorithms~\cite{SLT} to unleash the potential of SNNs. However, the inconsistency of SNNs across timesteps remains an unsolved issue that limits their performance, especially at low latency. Although~\cite{MPS} modifies the neural dynamics and guides the output between neighboring timesteps to alleviate this problem, further solutions that are more efficient and effective are still needed. To this end, this paper performs dual consistency optimization via the efficient stable spike, without any modification of neurons or architectures, and compatible with other improvements.

\textbf{Neuromorphic Object Recognition.} Unlike traditional static images, neuromorphic sensors such as event cameras represent data as sparse binary events with high temporal resolution and are able to capture motion information about the object~\cite{NDA,eventmix,SSNN}. The dimensionality of neuromorphic data can be denoted as $[t,x,y,p]$, where $t$ is the temporal index, $[x,y]$ are the spatial coordinates of the data, and $p$ is the polarity (positive polarity indicates that pixel brightness has increased above a certain threshold, and negative polarity is the opposite). This spatio-temporal property makes neuromorphic objects naturally suited for recognition by SNNs, and the power and speed benefits of combining the two have been widely demonstrated~\cite{yao2024spike,NDA}. There has been extensive work on optimizing SNNs for neuromorphic object recognition, but they still suffer from significant latencies~\cite{SLT,liu2025deeptage,SWformer,zhang2025staa} (typically more than 10 timesteps). In particular, \cite{MPS} points out that the temporal instability of SNNs is particularly pronounced in neuromorphic data and is a bottleneck limiting their low-latency performance. For these reasons, this paper focuses on enhancing neuromorphic object recognition with SNNs at ultra-low latencies. But notice that, as demonstrated in the experimental section, our method is also effective for static data.

\textbf{Consistency in SNNs.} Inspired by ANNs, previous methods have improved the spatial feature consistency of SNNs through knowledge distillation~\cite{KDSNN,guo2024enofsnn,BKDSNN} and contrastive learning~\cite{temporalcontrastivelearningspiking,zhang2024enhancing}. However, focusing solely on spatial aspects fails to leverage the spatio-temporal properties of SNNs fully, especially hindering their integration with neuromorphic data. To improve temporal consistency, especially in neuromorphic recognition performance, further studies distill the temporal dimension~\cite{TKS,TSSD,ding2025synergy} or promote consistency in representations between neuromorphic and static objects~\cite{he2024efficient}. However, while consistency has been facilitated, the lack of a stable anchor of consistency has limited the effectiveness of these methods. In this paper, we decouple the stable spikes that outline the feature skeleton as an anchor through the efficient ``AND" operation, thus guiding the variable spike maps towards consistency.

\vspace{-0.1in}
\section{Method}
We first describe the spiking neuron dynamics and the inconsistency problem in SNNs, and then detail how dual consistency optimization can be efficiently achieved by stable spike to improve the performance of SNNs.

\begin{figure*}
  \centering
  \includegraphics[width=0.95\linewidth]{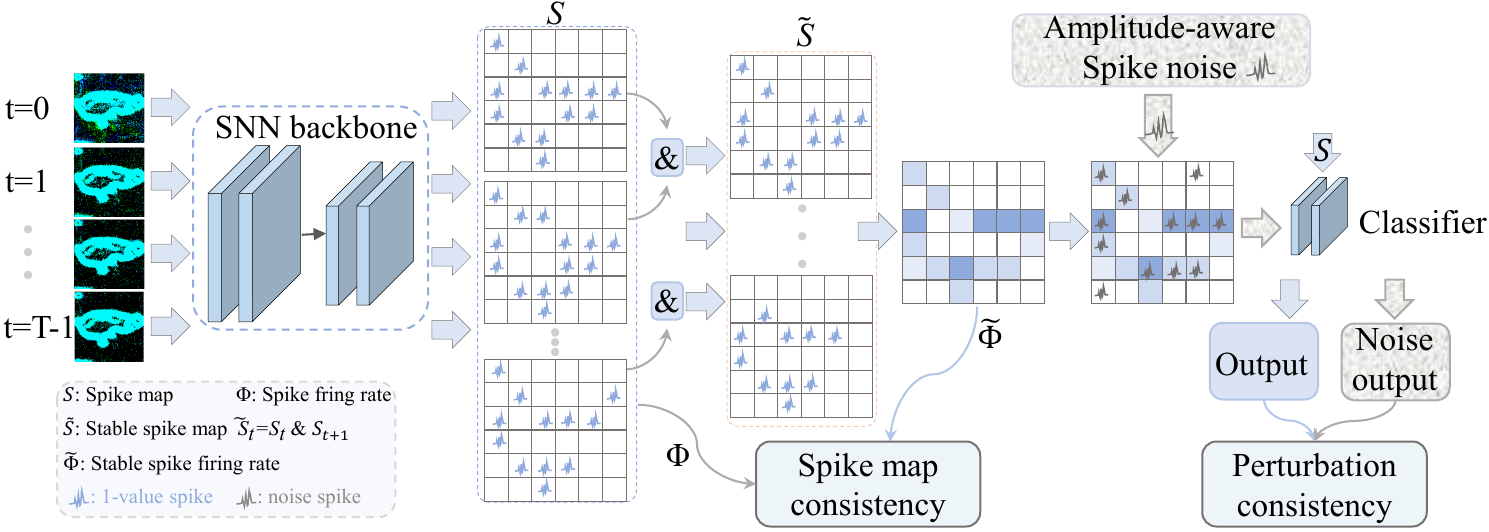}
  \vskip -0.1in
  \caption{The stable spikes decoupled by the minimal \& operation are used as the anchor. On the one hand, we promote the variable original spike maps to converge to the stable spike firing rate skeleton, i.e., spike map consistency. On the other hand, we introduce amplitude-aware spike noise to the stable spiking firing rate to preserve the key semantics and increase the feature diversity, allowing the SNN to be insensitive to the perturbation and promote generalization, i.e., perturbation consistency.}
  \vskip -0.2in
\end{figure*}

\subsection{Inconsistency Arising from Spike Dynamics}
The spiking neuron is the core element in SNNs that mimics the biological charge-firing mechanism to generate binary spike signals. In this paper, we use the most commonly used leaky integrate-and-fire (LIF) neuron model~\cite{MPS,STBP}, which balances biological plausibility with ease of implementation. The LIF neuron model iteratively undergoes three phases of charging, firing, and resetting within $T\ge1$ discrete timesteps. Upon receiving an input current $I$ from presynaptic neurons, the LIF neuron incorporates it into the membrane potential $H$, i.e., charging. The LIF neuron then determines whether the membrane potential exceeds a specific firing threshold $\vartheta$ to generate a spike $S$ or remain silent. During the resetting phase, the LIF neuron updates the membrane potential based on the fired spike, subtracting the membrane potential by the same amplitude as the threshold in a soft reset manner, and leaving it unchanged if no spike is fired. The iterative dynamics of LIF neurons within a single timestep can be expressed as:
\begin{equation}
H_{i,t}^{l}=(1-\frac{1}{\tau}) H_{i,t-1}^{l}+I_{i,t}^{l},
\label{eq1}
\end{equation}
\vskip -0.15in
\begin{equation}
S_{i,t}^{l} = \left\{
\begin{array}{cl}
1,\quad H_{i,t}^{l} \ge \vartheta \\
0,\quad H_{i,t}^{l} < \vartheta \\
\end{array},
\right.
\label{eq2}
\end{equation}
\vskip -0.15in
\begin{equation}
H_{i,t}^{l} = H_{i,t}^{l}-S_{i,t}^{l}\vartheta,
\label{eq3}
\end{equation}
where $l$, $i$, and $t$ denote the layer, neuron, and timestep indexes in the SNN, respectively. In particular, the membrane potential of LIF neurons leaks when the dynamics iterate to the next timestep, which is controlled by $\tau$ in Eq.~(\ref{eq1}) with a default value of 2.0.

The output of a spiking neuron at each timestep depends on both the input $I$ and its membrane potential value $H$, since the membrane potential is maintained across timesteps. The synergistic interaction of these two factors results in differences in the output of the SNN across timesteps. As shown in Fig.~\ref{fig1}, excessive differences between spike maps across timesteps lead to overall inconsistency, and plenty of highly variable spurious spikes impeding semantic feature extraction. In particular, since the membrane potential of the spiking neuron is usually initialized to 0, the output of the early timestep is more confusing compared to the late one~\cite{STBP,MPS}. The inconsistency across timesteps limits the overall performance of the SNN, especially when only early timesteps are available for low-latency inference.

\subsection{Stable Spike for Spike Map Consistency}

To promote multi-timestep consistency of SNNs, we separate the variable spike feature maps into consistent stable spike skeletons and unstable spurious redundant spikes. Therefore, we are able to train the spike feature maps to converge to a stable and consistent skeleton while ignoring the interference of redundant and variable spurious spikes.

Without loss of generality, we denote the spike feature map generated by the SNN over $T$ timesteps as $S \in \{0,1\}^{T \times B \times C \times H \times W}$, where $B$ is the batch size, $C$ is the number of channels, and $H$ and $W$ denote the height and width, respectively (temporarily ignore layer indices). To efficiently extract salient semantic features captured by the SNN over $T$ timesteps from the unstable $S$, we use the binary data-friendly ``AND" (\&) operation to compute the stable spike skeleton between adjacent spike maps. In particular, the \& operation is able to preserve 1-valued elements coexisting in the two binary data while eliminating unstable noise elements, and is naturally suited for modeling consistency in SNNs. Among the $T$ varying spike maps, we extract a total of $T-1$ stable spikes, which is expressed in mathematical form as follows:
\vskip -0.2in
\begin{equation}
\tilde{S}_{i,t} = S_{i,t}~\&~S_{i,t+1} = \left\{
\begin{array}{cl}
1,\quad &S_{i,t} = S_{i,t+1} = 1\\
0,\quad &else \\
\end{array},
\right.
\label{eq4}
\end{equation}
Fig.~\ref{fig1} visualizes the stable spikes decoupled by the \& operation, which reveal a clear feature skeleton with significantly less inconsistency compared to the original spike maps. 

Furthermore, we define the stable spike firing rate, $\tilde{\Phi}=\frac{1}{T-1}\sum_{t=0}^{T-2}\tilde{S}_t$, to represent the aggregated feature of stable spikes over multiple timesteps. $\tilde{\Phi}$ effectively represents the feature skeleton extracted by the SNN over multiple timesteps, so we align the variable spike map $S$ to it to improve consistency and ignore spurious spike interference. To this end, we take the stable spike firing rate $\tilde{\Phi}$ as an anchor point to guide the spike map $S_t$, and achieve the consistency constraint by backpropagation of the spike map consistency objective function.

During implementation, we use the mean squared error (MSE) function as the consistency guidance function and perform averaging on the original $T$ spike feature maps to obtain the spike fring rate $\Phi$. Thus, the mathematical form of spike map consistency alignment can be expressed as:
\vskip -0.2in
\begin{equation}
\begin{aligned}
\mathcal{L}_{spike} &= MSE(\tilde{\Phi},\Phi)\\
&= \frac{\sum_{i=1}^{C \times H \times W}(\frac{1}{T-1}\sum_{t=0}^{T-2}\tilde{S}_{i,t}-\frac{1}{T}\sum_{t=0}^{T-1}S_{i,t})^2}{C \times H \times W}
\end{aligned}
\label{eq5}
\end{equation}

\begin{table}
\caption{Alternative consistency function.}
\vskip -0.12in
\label{tab:loss}
 \tabcolsep=0.01\columnwidth
\centering
\scalebox{0.83}{
\begin{tabular}{c|c|ccc|ccc}
  \toprule
  \multirow{2}{*}{Dataset} & \multirow{2}{*}{Vanilla} & \multicolumn{3}{c|}{$\mathcal{L}_{spike}$} & \multicolumn{3}{c}{$\mathcal{L}_{noise}$}\\
  & & MSE & KL & Cosine & MSE & KL & Cosine \\ 
  \midrule
  CIFAR10-DVS & 72.9 & \textbf{77.1} & 76.2 & 75.6 & 76.1 & \textbf{77.1} & 76.0\\
  DVS-Gesture & 87.15 & \textbf{94.44} & 94.10 & 93.40 & 93.40 & \textbf{94.44} & 91.67\\
  \bottomrule
 \end{tabular}
}
\vskip -0.15in
\end{table}

During training, the gradient of $\mathcal{L}_{spike}$ directly contributes to the stabilization of the spike firing rate, while preventing performance degradation due to excessive variance over multiple timesteps. Compared to the indirect adjustment of membrane potential and final output~\cite{MPS}, the direct control of spike firing rate brings more significant consistency and superior discrimination performance. It is worth noting that our method is not limited to a specific consistency function. The results in Table~\ref{tab:loss} show that significant performance gains can also be achieved with KL divergence or cosine similarity, suggesting that our method promises to further unlock the performance of SNNs with an optimized consistency function implementation.

\subsection{Amplitude-Aware Spike Noise for Perturbation Consistency}

The feature diversity of neural networks can facilitate generalization under the premise of consistently preserving semantic information~\cite{Li_2021_ICCV,gao2022hyperbolic}. Inspired by this, we expect to further increase the spike feature diversity to promote the generalization of SNNs. However, while ANNs can benefit from vanilla Gaussian noise~\cite{Li_2021_ICCV}, this does not work for SNNs due to the following two spike-specific characteristics: (1) the inherent discrete property of binary spikes makes the addition of continuous Gaussian noise during training lead to training-inference precision mismatches; and (2) discrete spike firing rate is more sensitive to noise amplitude compared to floating-point numerical activation. Therefore, we propose amplitude-aware spike noise, which ensures that the SNN conveys discrete information through noise spikes during both training and inference, while the noise amplitude depends on the spike firing rate, ensuring that the perturbation promotes generalization without causing degradation.

A prerequisite for noise perturbation of features to facilitate generalization is the preservation of key semantic information~\cite{NEURIPS2019_15f99f21}. Since the stable spike firing rate $\tilde{\Phi}$ filters out differences across timesteps as a stable feature skeleton, we add noise to $\tilde{\Phi}$ instead of directly perturbing the original features as in an ANN~\cite{Li_2021_ICCV,NEURIPS2019_15f99f21}. The discrete value range of the stable spike firing rate $\tilde{\Phi}$ decoupled within $T$ timesteps is $\{0,\frac{1}{T-1},\cdots,\frac{T-2}{T-1},1\}$. To maintain the discrete property, we sample binary spike noise $\varepsilon \in \{0,1\}$ from the Bernoulli distribution with probability $p$ and add it to the stable spike firing rate $\tilde{\Phi}$ for perturbation. Specifically, the spike noise $\varepsilon$ is obtained by randomly sampling the value $p_{tmp}$ from $[0,1]$ and comparing it to the noise probability $p$. If $p_{tmp}$ is less than or equal to $p$, a spike is generated. This is similar to value-dependent rate encoding of inputs, but requires only one random number generation and comparison, whereas input encoding requires multiple random samples to approximate the Poisson distribution~\cite{10.1007/978-3-030-58526-6_24,9747906,guo2021neural}. After perturbation by the noise spike, the value range of the perturbed firing rate is $\{0,\frac{1}{T-1},\cdots,\frac{T-2}{T-1},1,1+\frac{1}{T-1},\cdots,1+\frac{T-2}{T-1},2\}$, which remains the discrete property.

\begin{figure*}[t]
  \centering
  \includegraphics[width=0.9\linewidth]{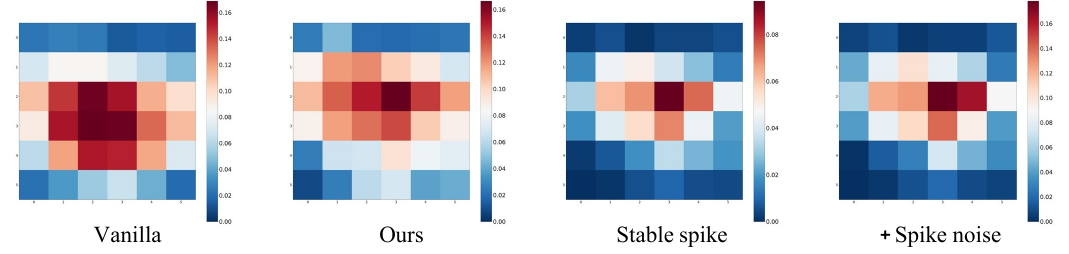}
  \vskip -0.13in
  \caption{Comparison of spike firing rates. Our spike firing rate shows a clearer feature profile compared to the vanilla SNN by converging to the stable spike and reducing spike interference. Additionally, the proposed amplitude-aware spike noise preserves the key semantic features of stable spikes while increasing feature diversity, thereby improving generalization. To reflect the implementation, we visualize the SNN backbone output of VGG-9 with a spike map size of $6 \times 6$.}
  \vskip -0.25in
\end{figure*}

On the one hand, if the spike feature value is small, excessive noise will disrupt the key semantics and cause performance degradation; on the other hand, a large spike feature value is insensitive to small noise perturbations, making it difficult to promote generalizability. To this end, we generate amplitude-aware spike noise by applying small perturbations to low firing rate elements and significant perturbations to higher firing rate elements, with the stable firing rate $\tilde{\Phi}$ as the reference. For the implementation, we take the firing rate value of the corresponding position element in $\tilde{\Phi}$ as the spike noise probability, i.e., $p_{c,i,j}=\tilde{\Phi}_{c,i,j}$, where $(c,i,j)$ denotes the position index. The amplitude-aware pulse noise can be expressed as:
\begin{equation}
\varepsilon_{c,i,j} = \left\{
\begin{array}{cl}
1,\quad &\text{with probability}~\tilde{\Phi}_{c,i,j}\\
0,\quad &else \\
\end{array}.
\right.
\end{equation}
In this way, elements with a higher stable spike firing rate are more likely to be perturbed by a 1-value spike, and vice versa, elements with low firing rate are prevented from being excessively perturbed.

After performing the noise perturbation on the stable spike firing rate to obtain $\Phi_{noise} = \tilde{\Phi} + \varepsilon$, we forward propagate the perturbed $\Phi_{noise}$ to generate the final prediction $O_{noise}$. To allow the SNN to learn generalized representations, we align the probability distributions of the noisy prediction $O_{noise}$ with the original clean prediction $O$. The temperature $\alpha$ softening output logit is first used to obtain its probability distribution:
\vskip -0.17in
\begin{equation}
p_j=\frac{e^{O_j/\alpha}}{\sum^K_{k=1}{e^{O_k/\alpha}}},p_{noise,j}=\frac{e^{O_{noise,j}/\alpha}}{\sum^K_{k=1}{e^{O_{noise,k}/\alpha}}},
\end{equation}
where $j$ denotes the $j$-th class, $K$ denotes a total of $K$ classes, and temperature $\alpha$ is set to 2. We use the KL divergence to promote consistency of the probability distribution after softening:
\vskip -0.17in
\begin{equation}
\mathcal{L}_{noise}=\alpha^2KL(O||O_{noise})=\alpha^2\sum^{K}_{k=1}{O_klog(\frac{O_k}{O_{noise,k}})}.
\label{eq7}
\end{equation}

\begin{algorithm}[t]
    \caption{The stable spike algorithm.}
    \label{alg: stablespike}
\begin{algorithmic}[1]
    \STATE {\bfseries Input:} training dataset $\{ x_i, {y_i}\}_{i=1}^B$, timestep $T$
    \STATE Initialization of parameters $\theta$ in the SNN $f(\cdot)$
    \FOR{$t=0$ {\bfseries to} $T-1$}
    \STATE Forward propagation calculates the backbone spike map $S_t=f(x)$
    \IF{$t>0$}
    \STATE Calculate the stable spike $\tilde{S}_{t-1}=S_{t-1}\&S_t$
    \ENDIF
    \STATE Forward propagation produces the output $O_t$
    \ENDFOR
    \STATE Adding amplitude-aware spike noise $\varepsilon$ to the stable spike firing rate $\Phi_{noise}=\tilde{\Phi}+\varepsilon$
    \STATE Calculate the noise output $O_{noise}$
    \STATE Calculate the dual consistency losses $\gets$ Eq.~\ref{eq8}
    \STATE Backpropagation to optimize parameters $\theta$
    \STATE {\bfseries Output:} Consistency-optimized SNN
\end{algorithmic}
\end{algorithm}

It is worth noting that the original SNN output $O$ averaged over $T$ timesteps is used as the final output; the noise prediction without temporal dimension is used directly as the final output. In addition, we compute stable spikes and perform consistency guidance only for the backbone features of the SNN, allowing additional forward propagation only past the classifier with negligible overhead. 

During training, the spike map consistency loss (Eq.~\ref{eq5}) and perturbation consistency loss (Eq.~\ref{eq7}) are seamlessly combined with the cross-entropy loss of the classification task by balancing coefficients $\beta$ and $\gamma$:
\begin{equation}
\mathcal{L}_{total}=\mathcal{L}_{CE} + \beta \mathcal{L}_{spike} + \gamma \mathcal{L}_{noise}.
\label{eq8}
\end{equation}
The complete algorithm for training SNNs by stable spike is shown in Algorithm~\ref{alg: stablespike}.

\begin{table}[t]
  \centering
  \tabcolsep=0.01\columnwidth
  \caption{Ablation results across different architectures.}
  \label{tab:ablation}
  \vskip -0.1in
  \scalebox{0.9}{
  \begin{tabular}{ccccc}
  \toprule
  Dataset & Method & VGG-9 & ResNet-18 & QKFormer\\
  \midrule
  \multirow{4}{*}{\shortstack{CIFAR10-\\DVS}} & Baseline & $72.9$ & $66.1$ & $81.2$ \\ & +$\mathcal{L}_{spike}$ & $75.2_{+2.4}$ & $69.7_{+3.6}$ & $82.5_{+1.3}$ \\ & +$\mathcal{L}_{noise}$ & $75.4_{+2.6}$ & $68.5_{+2.4}$ & $82.0_{+0.8}$\\
  \rowcolor{ourcolor} \cellcolor{white} &  +Both & $\textbf{77.1}_{+4.2}$ & $\textbf{70.3}_{+4.2}$ & $\textbf{82.9}_{+1.7}$ \\ 
  \hline
  \multirow{4}{*}{\shortstack{DVS-\\Gesture}} & Baseline & 87.15 & 81.59 & 93.75 \\ & + $\mathcal{L}_{spike}$ & $91.32_{+4.17}$ & $84.38_{+2.79}$ & $94.44_{+0.69}$\\ & +$\mathcal{L}_{noise}$ & $94.09_{+6.94}$ & $83.68_{+2.09}$ & $94.79_{+1.04}$\\
 \rowcolor{ourcolor} \cellcolor{white} &  +Both & $\textbf{94.44}_{+7.29}$ & $\textbf{85.42}_{+3.83}$ & $\textbf{95.49}_{+1.74}$\\
  \bottomrule
 \end{tabular}
  }
  \vskip -0.15in
\end{table}

\vspace{-0.05in}
\section{Experiments}
\vspace{-0.05in}
We focus on neuromorphic benchmarks CIFAR10-DVS~\cite{CIFAR10-DVS}, DVS-Gesture~\cite{DVS-Gesture}, and N-Caltech101~\cite{N-Caltech101}, while also conducting validation on static CIFAR10/100 and ImageNet using VGG, ResNet, and Transformer-style architectures to demonstrate the effectiveness and generalizability of our method. For VGG-9 and ResNet-18, we use the same training strategy as MPS~\cite{MPS}; for QKFormer, we also follow its original strategy~\cite{QKFormer}. If not specified, the consistency loss balance coefficients $\beta$ and $\gamma$ are set to 1.0, and the timestep is 4 to reflect the low-latency performance of the SNN. The detailed experimental setup can be found in the \textbf{Supplementary Material}.

\vspace{-0.05in}
\subsection{Ablation Study}
To demonstrate the effectiveness and versatility of the proposed method, we performed ablation studies with three architectures on CIFAR10-DVS and DVS-Gesture, and the results are shown in Table~\ref{tab:ablation}. The results show that the proposed method consistently delivers performance gains for different SNN architectures, in particular up to 7.29\% for DVS-Gesture recognition with VGG-9. When applied to the QKFormer~\cite{QKFormer} with SOTA performance, the performance is further improved, demonstrating the superior generalizability of our method.
\begin{table}[t]
  \centering
  \tabcolsep=0.01\columnwidth
  \caption{The effectiveness of amplitude adaption and discrete spike noise. $p$ is the fixed noise spike probability and $std$ is the standard deviation of the continuous Gaussian noise.}
  \label{tab:noise}
  \vskip -0.1in
  \scalebox{0.85}{
  \begin{tabular}{ccc|ccc}
  \toprule
  Dataset & Ablation & Acc. (\%) & Dataset & Ablation & Acc. (\%)\\
  \midrule
  \multirow{7}{*}{\shortstack{CIFAR10-\\DVS}} & $p=0.4$ & 74.8 & \multirow{7}{*}{\shortstack{DVS-\\Gesture}} & $p=0.4$ & 87.15\\
  & $p=0.5$ & 75.1 & & $p=0.5$ & 88.89\\
  & $p=0.6$ & 75.5 & & $p=0.6$ & 86.81\\
  \cmidrule{2-3}\cmidrule{5-6}
  & $std=0.1$ & 74.9 & & $std=0.1$ & 88.19 \\
  & $std=0.5$ & 75.3 & & $std=0.5$ & 91.67\\
  & $std=1.0$ & 74.5 & & $std=1.0$ & 89.93\\
  \cmidrule{2-3}\cmidrule{5-6}
  & \cellcolor{ourcolor} Ours & \cellcolor{ourcolor}\textbf{77.1} & & \cellcolor{ourcolor} Ours & \cellcolor{ourcolor}\textbf{94.44} \\
  \bottomrule
 \end{tabular}
  }
  \vskip -0.1in
\end{table}

In addition, we examine the effects of two key components of amplitude-aware spike noise: amplitude adaptation and spike noise. We perturbed the stable spike firing rate with a fixed noise spike probability $p$ and continuous Gaussian noise with mean value 0, respectively, and the results are shown in Table~\ref{tab:noise}. The results show that using the fixed noise spike probability whose amplitude is uncontrollable significantly degrades the performance of the proposed method; replacing the discrete spike noise with continuous Gaussian noise also leads to degraded performance. In particular, when the fixed noise probability is excessively high, leading to overly strong perturbations ($p=0.75$), the model fails to converge. In contrast, the proposed amplitude-aware spike noise considers the discrete, noise-sensitive properties of SNNs and demonstrates superior performance.

\begin{table}[t]
  \tabcolsep=0.08\columnwidth
  \centering
  \caption{Comparison (\%) of AND, OR, and XOR bit operations.}
  \vskip -0.1in
  \label{tab:OR}
  \scalebox{0.85}{
\begin{tabular}{ccccc}
  \toprule
  Dataset & AND & OR & XOR\\
  \midrule
  CIFAR10-DVS & \textbf{77.1} & 68.9 & 74.5\\
  DVS-Gesture & \textbf{94.44} & 88.54 & 89.58\\
  \bottomrule
 \end{tabular}
  }
  \vskip -0.22in
\end{table}

Table~\ref{tab:OR} compares the performance of AND and other bit operations (OR and XOR). The AND operation consistently retrieves (1, 1) spike pairs, whereas the OR and XOR operations retrieve the \{(1, 0), (0, 1), (1, 1)\} and \{(0, 1), (1, 0)\} patterns, respectively. The results show that the AND operation achieves optimal performance, which aligns with our insights on consistency. Notably, the OR operation caused significant degradation due to the simultaneous retrieval of consistent/inconsistent spike patterns triggering confusion.

\begin{figure}
    \centering
    \includegraphics[scale=0.16]{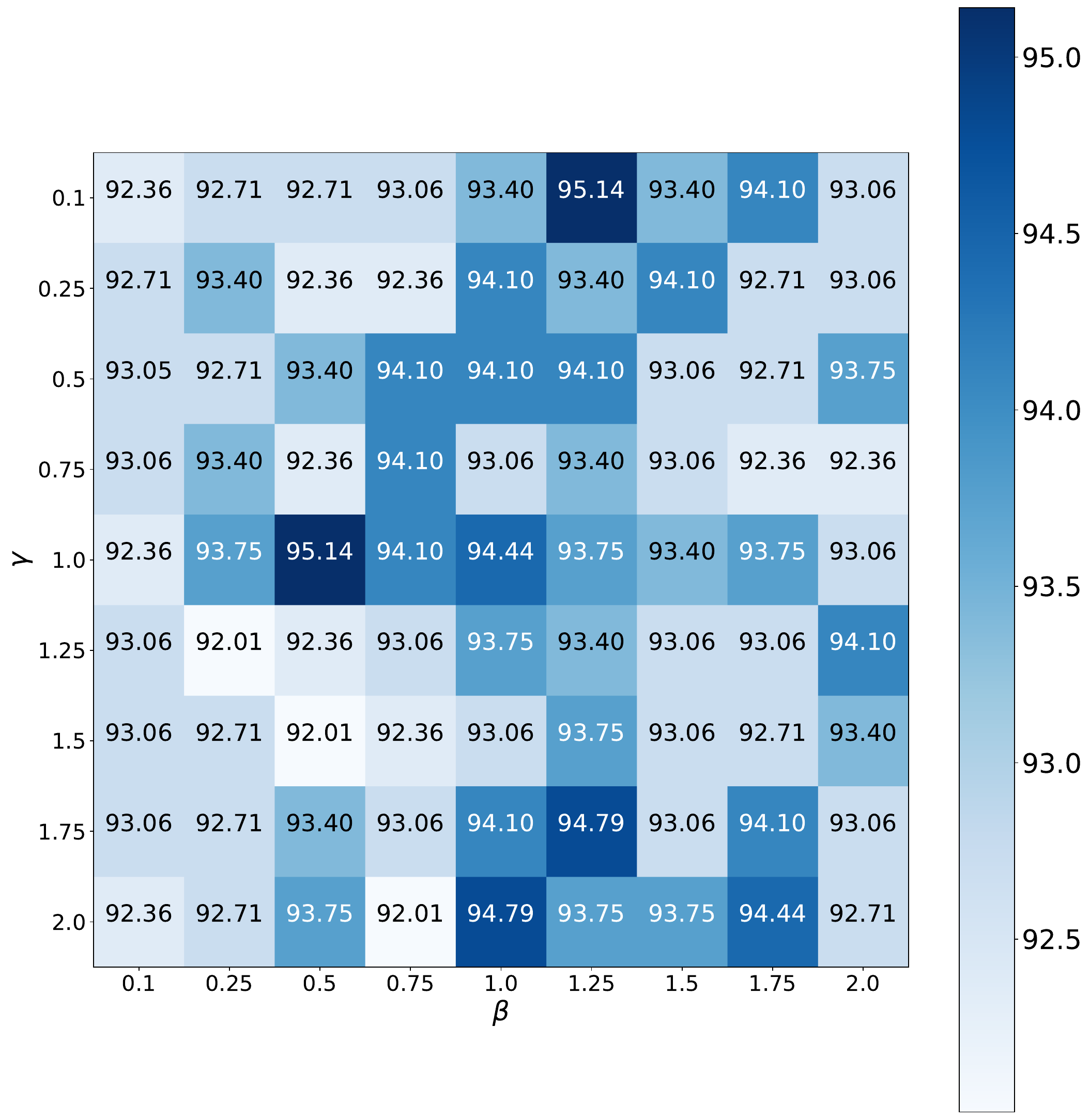}
    \vspace{-0.1in}
    \caption{Performance (\%) of different balance coefficients $\beta$ and $\gamma$ on DVS-Gesture.} 
    \label{fig:hyper}
     \vspace{-0.15in}
\end{figure}

To investigate the influence of $\beta$ and $\gamma$ on the performance, we conducted experiments on DVS-Gesture in the range $\{0.1,0.25,0.5,0.75,1.0,1.25,1.5,1.75,2.0\}$. The results in Fig.~\ref{fig:hyper} show that these coefficients affect the performance to a certain extent, where the lowest accuracy is 92.01\%, while the highest accuracy reaches 95.14\%. It is worth noting that although the performance fluctuates, the lower bound of our method still far exceeds the performance of the vanilla SNN (87.15\%). This suggests that our method can be plug-and-play to improve performance and offers higher potential after adjusting the balancing coefficients.

\begin{table}[t]
\centering
  \tabcolsep=0.05\columnwidth
  \caption{Comparative results (\%) on neuromorphic datasets. $\dag$ denotes knowledge transfer from static data.}
\vskip -0.12in
 \label{tab:com_neuro}
 \tabcolsep=0.008\columnwidth
 \begin{threeparttable}
 \scalebox{0.8}{
 \begin{tabular}{p{0.1in}lccc}
  \toprule
   & Method & Architecture & $T$ $\downarrow$ & Acc. (\%) $\uparrow$ \\
  \midrule
  \multirow{17}{*}{\rotatebox{90}{\textbf{CIFAR10-DVS}}}
  & STAA-SNN$^{CVPR'25}$~\citep{zhang2025staa} & VGG-13 & 16 & 82.1\\
   & SLT$^{AAAI'24}$~\citep{SLT} & VGGSNN & 10 & 81.46\\ 
   & DeepTAGE$^{ICLR'25}$~\citep{liu2025deeptage} & VGG-11 & 10 & 81.23\\
   & EnOF-SNN$^{NeurIPS'24}$~\citep{guo2024enofsnn} & ResNet20 & 10 & 80.5\\ 
   & BKDSNN$^{ECCV'24}$~\citep{BKDSNN} & Wide-7B-Net & 8 & 72.2\\ 
   & SSNN$^{AAAI'24}$~\citep{SSNN} & VGG-9 & 5 & 73.63\\
\cline{2-5}
  & \multirow{2}{*}{MPS$^{ICLR'25}$~\citep{MPS}} & VGG-9 & 5 & 76.77\\ 
  & & VGGSNN & 4 & 83.2\\
  \cline{2-5}
 & \cellcolor{ourcolor} & \cellcolor{ourcolor} VGG-9 & \cellcolor{ourcolor} 4 & \cellcolor{ourcolor} \textbf{77.1}\\
 & \cellcolor{ourcolor} \multirow{-2}{*}{\textbf{Ours}} & \cellcolor{ourcolor} VGGSNN & \cellcolor{ourcolor} 4 & \cellcolor{ourcolor} \textbf{83.7}\\
  \cline{2-5}
   & SNN-ViT$^{ICLR'25}$~\citep{wang2025spiking} & SNN-ViT & 16 & 82.3\\
   & SWformer$^{ECCV'24}$~\citep{SWformer} & SWformer & 10 & 82.9\\ 
   & SEMM$^{NeurIPS'24}$~\citep{SEMM} & Spikingformer & 10 & 80.7\\ 
   & SpikingResformer$^{CVPR'24}$~\citep{Shi_2024_CVPR} & SpikingResformer & 10 & 81.5\\
   & MPS$^{ICLR'25}$~\citep{MPS} & SpikingResformer & 5 & 80.6\\
   & QKFormer$^{NeurIPS'24}$~\citep{QKFormer} & QKFormer & 4 & 81.2\\ 
   \cline{2-5}
 \rowcolor{ourcolor} \cellcolor{white} & \textbf{Ours} & QKFormer & 4 & \textbf{82.9}\\ 
  \hline
  \multirow{13}{*}{\rotatebox{90}{\textbf{DVS-Gesture}}}
  & MPS$^{ICLR'25}$~\citep{MPS} & VGG-9 & 5 & 93.23\\
  & SSNN$^{AAAI'24}$~\citep{SSNN} & VGG-9 & 5 & 90.74\\ 
  & CLIF$^{ICML'24}$~\citep{CLIF} & VGG-9 & 4 & 89.58\\ 
  & TAB$^{ICLR'24}$~\citep{TAB} & VGG-9 & 4 & 87.50\\ 
  & SLT$^{AAAI'24}$~\citep{SLT} & VGG-9 & 4 & 88.19\\ 
  \cline{2-5}
  & \cellcolor{ourcolor} \textbf{Ours} &  \cellcolor{ourcolor} VGG-9 & \cellcolor{ourcolor} 4 & \cellcolor{ourcolor} \textbf{94.44}\\
  \cline{2-5}
    & QKFormer$^{NeurIPS'24}$~\citep{QKFormer} & QKFormer & 16 & 98.60\\
     & SpikingResformer$^{CVPR'24}$~\citep{Shi_2024_CVPR} & SpikingResformer & 16 & 98.60\\
   & SEMM$^{NeurIPS'24}$~\citep{SEMM} & Spikingformer & 10 & 96.88\\
   & MPS$^{ICLR'25}$~\citep{MPS} & SpikingResformer & 5 & 94.44\\
   & QKFormer$^{NeurIPS'24}$~\citep{QKFormer} & QKFormer & 4 & 93.75\\ 
  \cline{2-5}
  \rowcolor{ourcolor} \cellcolor{white} & & & 4 & \textbf{95.49}\\
  \rowcolor{ourcolor} \cellcolor{white} & \multirow{-2}{*}{\textbf{Ours}} & \multirow{-2}{*}{QKFormer}& 16 & \textbf{98.61}\\

  \hline
  \multirow{13}{*}{\rotatebox{90}{\textbf{N-Caltech101}}}
  &TCJA-TET-SNN$^{TNNLS'24}$~\citep{TCJA} & CombinedSNN & 14 & 82.50\\
  & SWformer$^{ECCV'24}$~\citep{SWformer} & SWformer & 10 & 88.45\\ 
  & IMP+TET-S$^{NeurIPS'24}$~\citep{shen2024rethinking} & VGGSNN & 10 & 85.01\\
  & TKS$^{IEEE~TAI'24}$~\citep{TKS} & VGGSNN & 10 & 84.10\\
  & EventMix$^{Inf. Sci.'23}$~\citep{eventmix} & ResNet-18 & 10 & 79.47\\ 
  &TIM$^{IJCAI'24}$~\citep{TIM} & Spikformer & 10 & 79.00\\
  & NDA$^{ECCV'22}$~\citep{NDA} & VGG-11 & 10 & 78.20\\ 
  & Knowledge-Transfer$^{AAAI'24}$~\citep{he2024efficient}& VGGSNN & 10 & 93.18\tnote{$\dag$}\\ 
  & SSNN$^{AAAI'24}$~\citep{SSNN} & VGG-9 & 5 & 77.97 \\
  \cline{2-5}
  & \multirow{2}{*}{MPS$^{ICLR'25}$~\citep{MPS}} & VGG-9 & 5 & 82.71\\ 
  & & VGGSNN & 10 & 93.68\tnote{$\dag$}\\
  \cline{2-5}
  \rowcolor{ourcolor} \cellcolor{white} & & VGG-9 & 4 & \textbf{83.92}\\
  \rowcolor{ourcolor} \cellcolor{white}& \multirow{-2}{*}{\textbf{Ours}} & VGGSNN & 10 & \textbf{94.25}\tnote{$\dag$}\\
  \bottomrule
 \end{tabular}
 }
 \end{threeparttable}
\vskip -0.22in
\end{table}

\vspace{-0.08in}
\subsection{Comparison with Other Methods}
\vspace{-0.02in}
\textbf{Neuromorphic Benchmark.} Table~\ref{tab:com_neuro} shows the comparative results on neuromorphic benchmarks. Without any data augmentation, our VGG-9 achieved a competitive accuracy of 77.1\% on CIFAR10-DVS with only 4 timesteps. Using standard data augmentation and the VGGSNN architecture, recognition accuracy reaches 83.7\%. On DVS-Gesture and N-Caltech101, our method also achieves significant performance advantages at low latency. In particular, our method can be used in conjunction with Knowledge-Transfer~\cite{he2024efficient} to achieve 94.25\% accuracy on N-Caltech101 at 10 timesteps, exceeding all other methods.

\textbf{Static Dataset.} Table~\ref{tab:com_imagenet} presents the comparative results on ImageNet. We achieved an accuracy of 70.59\% using ResNet-34 with four timesteps, surpassing other comparative methods. On CIFAR10 and CIFAR100, we achieved accuracies of 96.73\% and 82.29\%, respectively, once again outperforming other methods. See \textbf{Supplementary Material} for details.

\begin{table}[!t]
 \centering
 \caption{Comparative results with other methods on ImageNet.}
\label{tab:com_imagenet}
\vskip -0.1in
\tabcolsep=0.03\columnwidth
 \begin{threeparttable}
 \scalebox{0.85}{
 \begin{tabular}{lccc}
  \toprule
  Method  & Architecture & $T$ & Acc.(\%)\\
  \midrule
  RateBP~\cite{yu2024advancing}$^{NeurIPS'24}$ & ResNet-34 & 4 & 70.01\\
  TAB~\cite{TAB}$^{ICLR'24}$ & ResNet34 & 4 & 67.78\\
  Shortcut~\cite{guo2024take}$^{NeurIPS'24}$ & ResNet34 & 4 & 68.14\\
  FSTA-SNN~\cite{yu2025fsta}$^{AAAI'25}$ & ResNet34 & 4 & 70.23\\
  STAA-SNN~\cite{zhang2025staa}$^{CVPR'25}$ & ResNet34 & 4 & 70.40\\
  IMP+LTS~\cite{shen2024rethinking}$^{ICLR'25}$ & ResNet34 & 4 & 68.90\\
  SSCL~\cite{zhang2024enhancing}$^{AAAI'24}$ & ResNet34 & 4 & 66.78\\
  EnOF~\cite{guo2024enofsnn}$^{NeurIPS'24}$ &  ResNet34 & 4 & 67.40\\
  MPS~\cite{MPS}$^{ICLR'25}$ & ResNet-34 & 4 & 69.03\\
  Strong2Weak~\cite{ding2025synergy}$^{NeurIPS'25}$ & ResNet-34 & 4 & 70.53\\
  Weak2Strong~\cite{ding2025synergy}$^{NeurIPS'25}$ & ResNet-34 & 4 & 69.87\\
  \hline
  \rowcolor{ourcolor} \textbf{Ours} & ResNet-34 & 4 & \textbf{70.59}\\
  \bottomrule
 \end{tabular}
 }
 \end{threeparttable}
\vskip -0.2in
\end{table}

\begin{figure*}[t]
\centering
\begin{subfigure}[b]{0.25\textwidth}
    \centering
    \includegraphics[width=\linewidth]{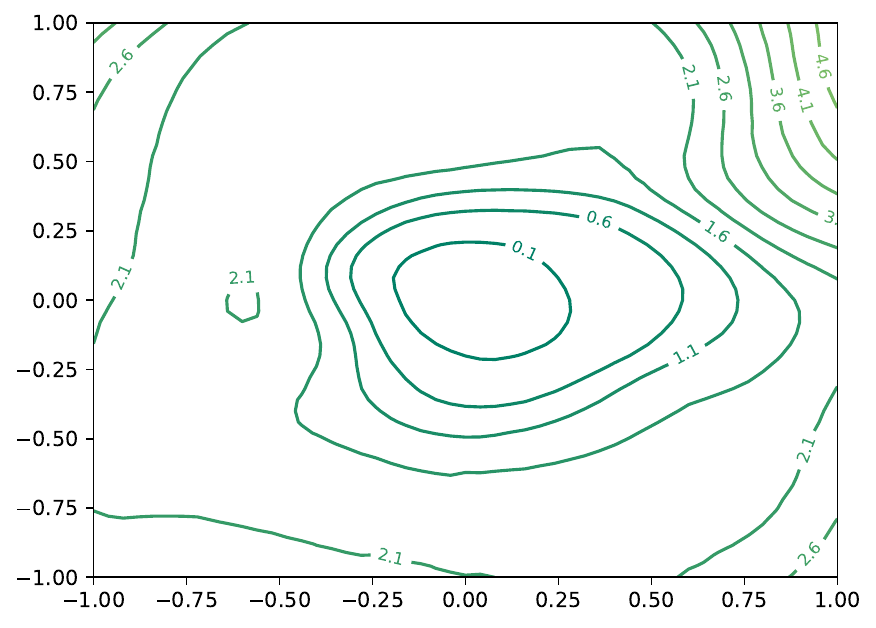}
    \caption{Vanilla}
  \end{subfigure}\hspace{-3mm}
  \begin{subfigure}[b]{0.25\textwidth}
    \centering
    \includegraphics[width=\linewidth]{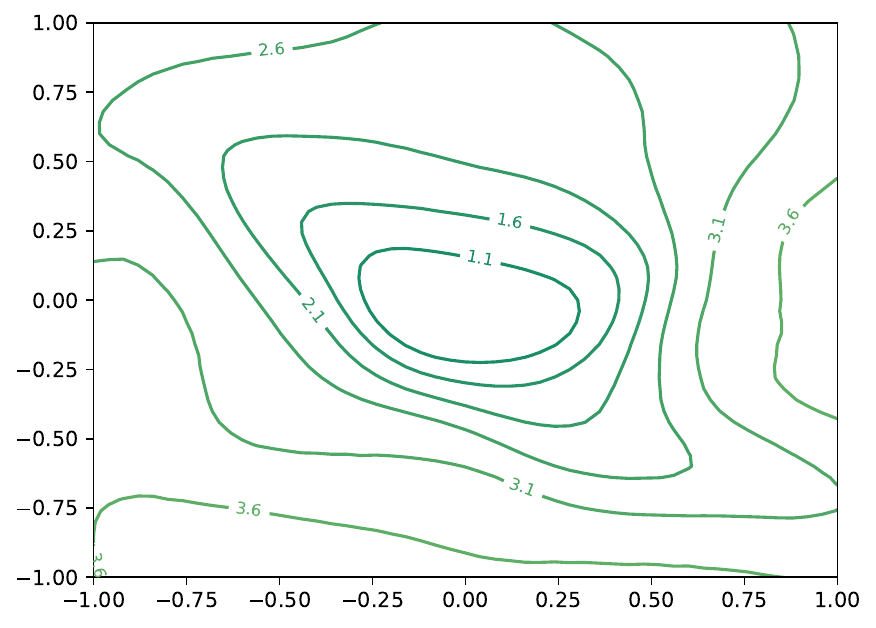}
    \caption{Ours}
  \end{subfigure}\hspace{-3mm}
  \begin{subfigure}[b]{0.25\textwidth}
    \centering
    \includegraphics[width=\linewidth]{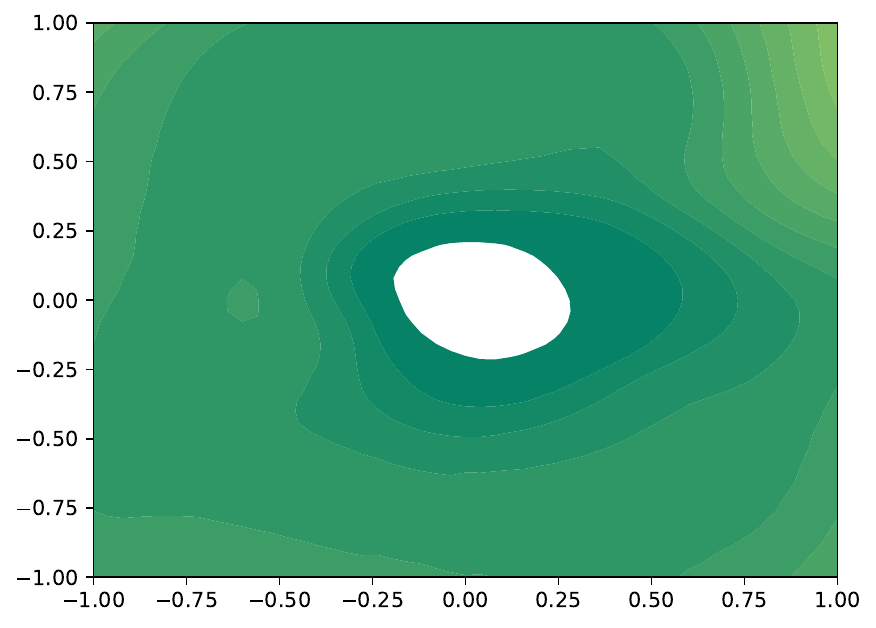}
    \caption{Vanilla}
  \end{subfigure}\hspace{-3mm}
  \begin{subfigure}[b]{0.25\textwidth}
    \centering
    \includegraphics[width=\linewidth]{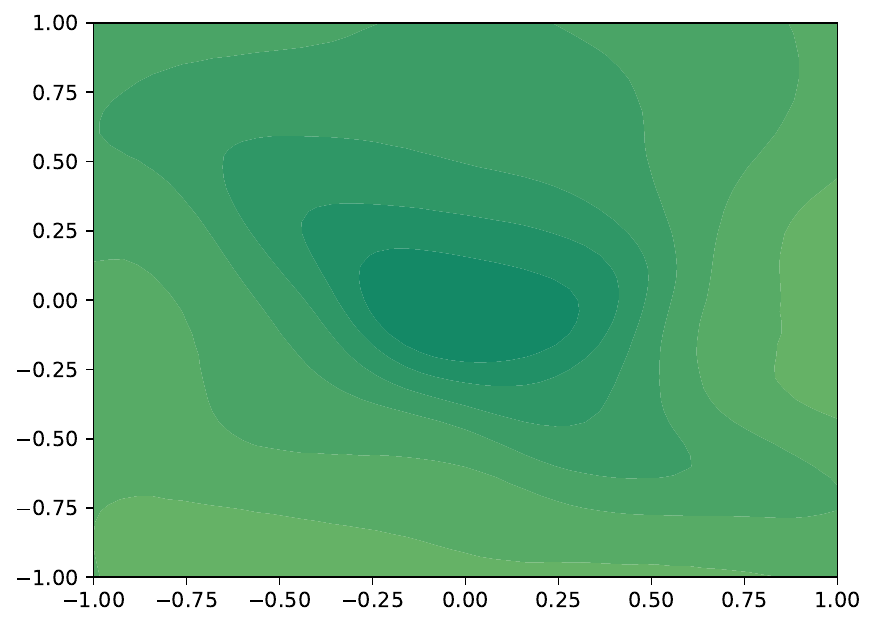}
    \caption{Ours}
  \end{subfigure}
\vskip -0.1in
\caption{Visualization of loss landscapes. (a) (c) The loss landscape of the vanilla SNN exhibits multiple local minima and saddle points. This complexity makes optimization susceptible to falling into local optima. (b) (d) Our method has a smoother and more centralized loss landscape with a clear global minimum, which makes optimization stable and convergent.}
\label{fig:loss}
\vskip -0.15in
\end{figure*}

\begin{table*}[t]
\centering
  \tabcolsep=0.02\columnwidth
  \caption{Scalable performance across timesteps. Our method improves performance consistently from ultra-low to high latencies.}
 \label{tab:timestep}
 \vskip -0.12in
  \scalebox{0.85}{
\begin{tabular}{c|cc|cc|cc|cc|cc}
  \toprule
\multirow{2}{*}{Dataset} & \multicolumn{2}{c|}{$T=2$} & \multicolumn{2}{c|}{$T=4$} & \multicolumn{2}{c|}{$T=6$} & \multicolumn{2}{c|}{$T=8$} & \multicolumn{2}{c}{$T=10$}\\
  & Vanilla & Ours & Vanilla & Ours & Vanilla & Ours & Vanilla & Ours & Vanilla & Ours \\ 
  \midrule
  CIFAR10-DVS & 72.6 & \cellcolor{ourcolor}$\textbf{74.6}_{+2.0}$ & 72.9 & \cellcolor{ourcolor}$\textbf{77.1}_{+4.2}$ & 73.7 & \cellcolor{ourcolor}$\textbf{77.5}_{+3.8}$ & 75.2 & \cellcolor{ourcolor}$\textbf{77.8}_{+2.6}$ & 75.5 & \cellcolor{ourcolor}$\textbf{78.2}_{+2.7}$\\
  DVS-Gesture & 83.68 & \cellcolor{ourcolor}$\textbf{92.01}_{+8.33}$ & 87.15 & \cellcolor{ourcolor}$\textbf{94.44}_{+7.29}$ & 88.19 & \cellcolor{ourcolor}$\textbf{94.10}_{+5.91}$ & 90.97 & \cellcolor{ourcolor}$\textbf{95.49}_{+4.52}$ & 92.01 & \cellcolor{ourcolor}$\textbf{95.49}_{+3.48}$\\
  \bottomrule
 \end{tabular}
  }
  \vskip -0.2in
\end{table*}

\begin{table}[t]
\centering
  \tabcolsep=0.015\columnwidth
  \caption{Comparison of spike firing rate (\%) and power consumption ($\times 10^6pJ$) on CIFAR10-DVS with VGG-9.}
 \label{tab:com_power}
 \vskip -0.1in
  \scalebox{0.8}{
\begin{tabular}{l|cccccccc|c}
  \toprule
  Layer $\rightarrow$ & 1 & 2 & 3 & 4 & 5 & 6 & 7 & 8 & Power $\downarrow$ \\
  \midrule
  Vanilla & \textbf{9.24} & 4.26 & 3.38 & 3.12 & 2.17 & 1.45 &  1.20 & 7.63 & 189.83\\
  \rowcolor{ourcolor} \textbf{Ours} & 10.14 & \textbf{3.61} & \textbf{3.20} & \textbf{2.88} & \textbf{1.70} & \textbf{1.25} & \textbf{1.10} & \textbf{9.81} & \textbf{181.02}\\
  \bottomrule
 \end{tabular}
  }
  \vskip -0.15in
\end{table}

\vskip -0.08in
\subsection{Loss Landscape Visualization}
\vspace{-0.02in}
To demonstrate the effectiveness of the proposed method from an optimization standpoint, Fig.~\ref{fig:loss} visually compares the loss landscapes of the vanilla SNN and the proposed method. In Fig.~\ref{fig:loss}(a)(c), the loss distribution of the vanilla SNN is dispersed and irregular, especially the upper-right and the center regions differ significantly. Additionally, local minima interfere with optimization, causing the model to be highly susceptible to falling into local optimal solutions. In contrast, the loss landscape of our method (Fig.~\ref{fig:loss}(b)(d)) is much smoother overall, without sharp variance despite the presence of spike noise perturbation during training. Consequently, our method is capable of converging to the global optimal solution with enhanced stability and efficiency, thereby facilitating generalization.

\vspace{-0.07in}
\subsection{Scalability with Timesteps}
\vspace{-0.03in}
Table~\ref{tab:timestep} shows the performance variation of the proposed method with different timesteps. We vary the timestep from the commonly used high latency ($T = 10$) to the ultra-low latency ($T = 2$). The results show that our method yields consistent performance gains that generalize to a wide range of latencies. In particular, the accuracy on DVS-Gesture is improved by 8.33\% at $T=2$, which significantly contributes to the development of ultra-low latency SNNs.

\vspace{-0.07in}
\subsection{Spike Firing Rate and Power Consumption}
\vspace{-0.03in}
To investigate the influence of our method on spike firing rate and power consumption, we present the comparative results with the vanilla SNN in Table~\ref{tab:com_power}. The results show that our method achieves lower firing rates and lower overall power consumption across all layers except the first. This indicates that our method not only improves the performance of SNNs, but also slightly reduces their power consumption during deployment.

\subsection{Compatibility with Other Methods}
\vskip -0.05in

\begin{table}[!t]
 \centering
 \caption{Our method can be seamlessly integrated with other methods and exhibits superior generalizability.}
\label{tab:generalization}
\vskip -0.12in
\tabcolsep=0.02\columnwidth
 \begin{threeparttable}
 \scalebox{0.85}{
\begin{tabular}{ccccc}
  \toprule
  Dataset & Type &  Method & Acc. (\%) & + Ours\\
  \midrule
  \multirow{3}{*}{\shortstack{CIFAR10-\\DVS}} & Spiking neuron & CLIF~\cite{CLIF} & 74.3 & \cellcolor{ourcolor}$\textbf{76.2}_{+1.9}$ \\
  & BN Layer & TAB~\cite{TAB} & 73.1 & \cellcolor{ourcolor}$\textbf{75.4}_{+2.3}$ \\
  & Training & SLT~\cite{SLT} & 74.1 & \cellcolor{ourcolor}$\textbf{75.9}_{+1.8}$\\
  \hline
  \multirow{3}{*}{\shortstack{DVS-\\Gesture}} & Spiking neuron & CLIF~\cite{CLIF} & 89.58 & \cellcolor{ourcolor}$\textbf{95.83}_{+6.25}$ \\
  & BN Layer & TAB~\cite{TAB} & 87.50 & \cellcolor{ourcolor}$\textbf{92.36}_{+4.86}$ \\
  & Training & SLT~\cite{SLT} & 88.19 & \cellcolor{ourcolor}$\textbf{90.97}_{+2.78}$\\
  \bottomrule
 \end{tabular}
 }
 \end{threeparttable}
\vskip -0.2in
\end{table}

Our method is a plug-and-play SNN enhancement algorithm that requires no modification to the neuron model or architecture. To further validate the compatibility, we combine our method with other SNN methods, including the improved neuron model CLIF~\cite{CLIF}, the SNN-oriented BN layer TAB~\cite{TAB}, and the efficient training algorithm SLT~\cite{SLT}. The results in Table~\ref{tab:generalization} show that our method consistently improves the performance of these methods, suggesting that our method has the potential to efficiently facilitate the performance of more SNN methods in the future.

\vspace{-0.07in}
\section{Conclusion}
\vspace{-0.03in}
In this paper, we decouple the stable spike skeleton via the ``AND" bit operation, and motivate the spike maps to converge towards it for consistency. Furthermore, we inject amplitude-aware spike noise to facilitate spike feature diversity while preserving key semantics, encouraging the SNN to generate perturbation-consistent predictions for generalization. Our method is neuron-architecture agnostic and can therefore be used to boost the performance of other SNN methods as a plug-and-play solution. Extensive experiments demonstrate the effectiveness and performance advantages of our method. We expect this work to inspire advancements in ultra-low-latency, high-performance SNNs.

\vskip -0.07in
\section*{Acknowledgments}
\vskip -0.03in
This work was supported by the National Natural Science Foundation of China under Grant No. 62276054.

{
    \small
    \bibliographystyle{ieeenat_fullname}
    \bibliography{main}
}

\appendix
\newpage
\section*{Description of the Supplementary Material}
This supplementary material contains four sections:

Section~\ref{AND_MI} interprets the effectiveness of the AND operation from the perspective of mutual information.

Section~\ref{detail} details the experimental setup, including the dataset and training strategy. 

Section~\ref{add_exp} presents additional experimental results, including further comparative experiments and extensions of our method. 

Section~\ref{overhead} provides an overhead analysis showing that our method requires only negligible training overhead and leaves inference completely unaffected.

Section~\ref{add_vis} provides supplementary visualizations demonstrating that the AND operation consistently extracts stable pulse skeletons.

\newpage

\section{Interpreting the AND Operation from the Perspective of Mutual Information}
\label{AND_MI}
The AND operation extracts the mutual information between adjacent timesteps. The $(1, 1)$ pair represents the consensus feature between two consecutive timesteps, which eliminates transient noise. According to the information bottleneck principle, the convergence of spike maps toward stable spikes maximizes mutual information in order to identify the minimal sufficient representation. In contrast, the XOR and XOR operations are inferior to the AND operation and may even exhibit degradation due to interference from (0,0) and (0,1).

\section{Experimental Details}
\label{detail}

\begin{table*}[!t]
 \centering
 \caption{Structures of VGG-9, VGGSNN, ResNet-18, and ResNet-19, where AP, GAP, and FC denote average pooling, global average pooling, and fully connected layers, respectively.}
 \label{model}
 \vskip -0.1in
 \begin{tabular}{ccccc}
  \toprule
  Stage & VGG-9 & VGGSNN & ResNet-18 & ResNet-19\\
  \midrule
  1 & - & -& Conv($3 \times 3@64$) & Conv($3 \times 3@128$)\\
  \hline
  1  &  \makecell{Conv($3 \times 3$@64) \\ Conv($3 \times 3$@128)} & \makecell{Conv($3 \times 3$@64) \\ Conv($3 \times 3$@128)} &
  \makecell{
  $\left(
 	    \begin{array}{cc}  
 			 \makecell{\text{Conv}(3 \times 3@64) \\ \text{Conv}(3 \times 3@64)}
        \end{array}
    \right)\times 2
  $}
  &
  \makecell{
  $\left(
 	    \begin{array}{cc}  
 			 \makecell{\text{Conv}(3 \times 3@128) \\ \text{Conv}(3 \times 3@128)}
        \end{array}
    \right)\times 3
  $}
  \\
  \hline
    & AP(stride=2) & AP(stride=2) &  - & -\\
  \hline
  2  & \makecell{Conv($3 \times 3$@256) \\ Conv($3 \times 3$@256)}  & \makecell{Conv($3 \times 3$@256) \\ Conv($3 \times 3$@256)} & 
  $\left(
 	    \begin{array}{cc}  
 			 \makecell{\text{Conv}(3 \times 3@128) \\ \text{Conv}(3 \times 3@128)}
        \end{array}
    \right)\times 2
  $
  &
  $\left(
 	    \begin{array}{cc}  
 			 \makecell{\text{Conv}(3 \times 3@256) \\ \text{Conv}(3 \times 3@256)}
        \end{array}
    \right)\times 3
  $
  
  \\
  \hline
    & AP(stride=2) & AP(stride=2) & - & -\\
  \hline
  3  & \makecell{Conv($3 \times 3$@512) \\ Conv($3 \times 3$@512)}  & \makecell{Conv($3 \times 3$@512) \\ Conv($3 \times 3$@512)} & $\left(
 	\begin{array}{cc}  
 			 \makecell{\text{Conv}(3 \times 3@256) \\ \text{Conv}(3 \times 3@256)}
 \end{array}
 \right)\times 2$
 &
 $\left(
 	\begin{array}{cc}  
 			 \makecell{\text{Conv}(3 \times 3@512) \\ \text{Conv}(3 \times 3@512)}
 \end{array}
 \right)\times 2$
 \\
  \hline
    & AP(stride=2) & AP(stride=2) & -\\
  \hline
  4  & \makecell{Conv($3 \times 3$@512) \\ Conv($3 \times 3$@512)} &   \makecell{Conv($3 \times 3$@512) \\ Conv($3 \times 3$@512)} & $\left(
 	\begin{array}{cc}  
 			 \makecell{\text{Conv}(3 \times 3@512) \\ \text{Conv}(3 \times 3@512)}
 \end{array}
 \right)\times 2$\\
  \hline
    & GAP,FC & AP,FC & GAP,FC & GAP,FC$\times 2$\\
  \bottomrule
 \end{tabular}
\end{table*}

\begin{table*}[!t]
 \centering
 \caption{Comparative results (\%) with adaptive Gaussian noise.}
 \label{com_adaptive}
 \vskip -0.1in
 \begin{threeparttable}
 \begin{tabular}{lcc}
  \toprule
   Method & CIFAR10-DVS & DVS-Gesture \\
  \midrule
  Adaptive Gaussian noise & 73.9 & 88.54\\ 
  \rowcolor{ourcolor} \textbf{Ours (Amplitude-aware spike noise)} & \textbf{77.1} & \textbf{94.44} \\ 
  \bottomrule
 \end{tabular}
 \end{threeparttable}
\end{table*}

\subsection{Neuromorphic Datasets.} We conducted experiments on three neuromorphic object recognition benchmark datasets: CIFAR10-DVS~\cite{CIFAR10-DVS}, DVS-Gesture~\cite{DVS-Gesture}, and N-Caltech101~\cite{N-Caltech101}. The CIFAR10-DVS~\cite{CIFAR10-DVS} dataset contains a total of 10,000 samples from ten categories. The DVS-Gesture~\cite{DVS-Gesture} dataset contains neuromorphic data for 11 hand gestures with 1176 training samples and 288 test samples. The spatial resolution of each sample in CIFAR10-DVS and DVS-Gesture is $128\times128$, which we downsampled to $48\times48$ and input to SNN, the standard input size for the community~\cite{MPS,QKFormer}. The N-Caltech101~\cite{N-Caltech101} dataset contains 8,109 samples, categorized into 101 categories, with an original resolution of $180 \times 240$. We also downsampled the samples in N-Caltech101 to $48\times48$.

We conducted our experiments on an Ubuntu 20.04.5 operating system with an NVIDIA 4090 GPU. To validate the versatility of the proposed method, we conducted experiments using various architectures, including VGG-9, ResNet-18, VGGSNN, and QKFormer. For VGG-9 and ResNet-18, we use the same training strategy as in~\cite{MPS}. The model was trained for 100 epochs using a stochastic gradient descent optimizer. The initial learning rate was set to 0.1, and it was scaled down tenfold every 30 epochs. The batch size is 64, and the weight decay is 1e-3. In particular, for VGG-9 and ResNet-18, we do not use any data augmentation during training. The firing threshold of the spiking neuron was set to 1.0, and the membrane potential time constant was set to 2.0. In the training of the SNN, the rectangular surrogate gradient function is employed for backpropagation, consistent with~\cite{MPS}. Table~\ref{model} shows the VGG-9 and ResNet-18 architectures.

When using VGGSNN for recognizing CIFAR10-DVS, we used random cropping and horizontal flipping as data augmentation. When using the VGGSNN architecture on the N-Caltech101 dataset, we incorporate the knowledge transfer strategy~\cite{he2024efficient} and employ the same training strategy as in~\cite{he2024efficient}. Similarly, when using the QKFormer architecture, we follow the training strategy outlined in the original paper~\cite{QKFormer} to ensure the fairness of the experiments. We set the balancing coefficients $\beta$ and $\gamma$ to 1.0 by default, and the experiments in Section 4.1 demonstrate the influence of these coefficients on performance. However, when we use VGGSNN to recognize the N-Caltech101 dataset and incorporate knowledge transfer, the magnitude of our loss does not match the original transfer loss. At this point, we set the balancing coefficients to 1e-3.

\begin{table*}[!t]
 \centering
 \caption{Comparative results (\%) under extended training epochs.}
 \label{com_epoch}
 \vskip -0.1in
 \begin{threeparttable}
 \begin{tabular}{cllcc}
  \toprule
  Dataset & Method & 100 Epoch & 200 Epoch & 300 Epoch \\
  \midrule
  \multirow{2}{*}{CIFAR10-DVS} 
  & Baseline & 72.9 & 74.1 & 74.2\\ 
  & \cellcolor{ourcolor}\textbf{Ours} & \cellcolor{ourcolor}\textbf{77.1} & \cellcolor{ourcolor}\textbf{77.2} & \cellcolor{ourcolor}\textbf{77.5}\\ 
  \hline
  \multirow{2}{*}{DVS-Gesture} 
  & Baseline & 87.15 & 89.58 & 89.58\\ 
  & \cellcolor{ourcolor} \textbf{Ours} & \cellcolor{ourcolor} \textbf{94.44} & \cellcolor{ourcolor} \textbf{95.14} & \cellcolor{ourcolor} \textbf{95.49}\\ 
  \bottomrule
 \end{tabular}
 \end{threeparttable}
\end{table*}

\subsection{Static Datasets.} For static images, we conducted experiments on the CIFAR10 and CIFAR100 datasets. Both the CIFAR-10 and CIFAR-100 datasets contain 60,000 $32 \times 32$ images, 50,000 of which are in the training set and 10,000 of which are in the test set. We normalized the CIFAR10 and CIFAR100 samples to a zero mean and unit variance. Then, we applied the standard data augmentation strategies, AutoAugment~\cite{Autoaugment} and Cutout~\cite{cutout}.

For CIFAR10 and CIFAR100, we applied the ResNet-18 and ResNet-19 architectures. We used a stochastic gradient descent optimizer with a momentum of 0.9 to train for 300 epochs. The initial learning rate was set to 0.1, and we employed a cosine annealing learning rate strategy. The batch size is 128, and weight deacy is 5e-4. Additionally, we use the QKFormer architecture to recognize static targets with the same training strategy as the original~\cite{QKFormer}.

On the ImageNet dataset, we train 300 epochs using ResNet-18 with a batch size of 256. The initial learning rate value is 0.1 and the learning rate is adjusted using a cosine annealing strategy. The default input size is $224\times224$, which is consistent with other methods.

Table~\ref{model} illustrates the architectures of the VGG-9, VGGSNN, ResNet-18, and ResNet-19 models that were used in the experiment. We use the same architecture and experimental setup as in our method when we reproduce CLIF~\cite{CLIF}, TAB~\cite{TAB}, and SLT~\cite{SLT}, but we use the officially released core code implementation to ensure performance.

\begin{table*}[!t]
 \centering
 \caption{Further comparative results under identical settings as MPS~\cite{MPS}.}
 \label{com_MPS}
 \vskip -0.1in
 \begin{threeparttable}
 \begin{tabular}{cllcc}
  \toprule
  Dataset & Method & Architecture & $T \downarrow$ & Accuracy (\%) $\uparrow$ \\
  \midrule
  \multirow{4}{*}{CIFAR10-DVS} 
  & MPS$^{ICLR'25}$~\citep{MPS} & VGG-9 & 5 & 76.77\\ 
  & \cellcolor{ourcolor} \textbf{Ours} & \cellcolor{ourcolor} VGG-9 & \cellcolor{ourcolor} 5 & \cellcolor{ourcolor} \textbf{77.20}\\ 
  \cline{2-5}
   & MPS$^{ICLR'25}$~\citep{MPS} & VGGSNN & 4 & 83.20\\ 
  & \cellcolor{ourcolor} \textbf{Ours} & \cellcolor{ourcolor} VGGSNN & \cellcolor{ourcolor} 4 & \cellcolor{ourcolor} \textbf{83.70}\\ 
  \hline
  \multirow{2}{*}{DVS-Gesture} 
  & MPS$^{ICLR'25}$~\citep{MPS} & VGG-9 & 5 & 93.23\\ 
  & \cellcolor{ourcolor} \textbf{Ours} & \cellcolor{ourcolor} VGG-9 & \cellcolor{ourcolor} 5 & \cellcolor{ourcolor} \textbf{95.14}\\ 
  \hline
  \multirow{4}{*}{N-Caltech101} 
  & MPS$^{ICLR'25}$~\citep{MPS} & VGG-9 & 5 & 82.71\\ 
  & \cellcolor{ourcolor} \textbf{Ours} & \cellcolor{ourcolor} VGG-9 & \cellcolor{ourcolor} 5 & \cellcolor{ourcolor} \textbf{84.03}\\ 
  \cline{2-5}
   & MPS$^{ICLR'25}$~\citep{MPS} & VGGSNN & 10 & 93.68\\ 
  & \cellcolor{ourcolor} \textbf{Ours} & \cellcolor{ourcolor} VGGSNN & \cellcolor{ourcolor} 10 & \cellcolor{ourcolor} \textbf{94.25}\\ 
  \bottomrule
 \end{tabular}
 \end{threeparttable}
\end{table*}

\begin{table*}[!t]
 \centering
 \caption{Computing consistency loss by applying the AND operation across additional layers can further enhance performance.}
 \label{com_dense}
 \vskip -0.1in
 \begin{threeparttable}
 \begin{tabular}{cllcc}
  \toprule
  Dataset & Method & Architecture & $T \downarrow$ & Accuracy (\%) $\uparrow$ \\
  \midrule
  \multirow{2}{*}{CIFAR10-DVS} 
  & Baseline & VGG-9 &  4 & 72.9 \\ 
  & \cellcolor{ourcolor} \textbf{Ours}  & \cellcolor{ourcolor} VGG-9 &  \cellcolor{ourcolor} 4 & \cellcolor{ourcolor} \textbf{77.1}\\ 
   & \cellcolor{ourcolor} \textbf{Ours Dense}  & \cellcolor{ourcolor} VGG-9 &  \cellcolor{ourcolor} 4 & \cellcolor{ourcolor} \textbf{78.8}\\ 
  \hline
  \multirow{2}{*}{DVS-Gesture} 
  & Baseline & VGG-9 &  4 & 87.15 \\ 
  & \cellcolor{ourcolor} \textbf{Ours} & \cellcolor{ourcolor} VGG-9 &  \cellcolor{ourcolor} 4 & \cellcolor{ourcolor} \textbf{94.44} \\ 
  & \cellcolor{ourcolor} \textbf{Ours Dense} & \cellcolor{ourcolor} VGG-9 &  \cellcolor{ourcolor} 4 & \cellcolor{ourcolor} \textbf{95.49} \\ 
  \bottomrule
 \end{tabular}
 \end{threeparttable}
\end{table*}

\section{Additional Experimental Results}
\label{add_exp}

\subsection{Comparison with Adaptive Gaussian Noise}

To further demonstrate that discrete spike noise outperforms continuous Gaussian noise in SNNs, we construct adaptive Gaussian noise using the spike firing rate as its standard deviation. As shown in Table~\ref{com_adaptive}, our amplitude-aware spike noise still exhibits significant performance advantages over adaptive Gaussian noise.

\subsection{Further Comparison with the Baseline}

The VGG-9 architecture was trained for 100 epochs by default. To show that our method consistently outperforms the baseline model rather than due to underfitting, we extended the training cycle to 300 epochs. The comparative results are presented in Table~\ref{com_epoch}. These results demonstrate that our method consistently yields significant performance gains.

\begin{table*}[!t]
 \centering
 \caption{Comparative results on static CIFAR10 and CIFAR100 datasets.}
 \label{com_cifar}
 \vskip -0.1in
 \begin{threeparttable}
 \begin{tabular}{cllcc}
  \toprule
  Dataset & Method & Architecture & $T \downarrow$ & Accuracy (\%) $\uparrow$ \\
  \midrule
  \multirow{12}{*}{CIFAR10} 
  & QKFormer$^{NeurIPS'24}$~\citep{QKFormer} & QKFormer & 4 & 96.18\\ 
  & SNN-ViT$^{ICLR'25}$~\citep{wang2025spiking} & SNN-ViT & 4 & 96.10\\
 & RateBP$^{NeurIPS'24}$~\cite{yu2024advancing} &  ResNet-19 & 4 & 96.26 \\
 & DeepTAGE$^{ICLR'25}$~\citep{liu2025deeptage} & ResNet-18 & 4 & 95.86\\
   & SLT$^{AAAI'24}$~\citep{SLT} &  ResNet-19 & 4 & 95.18 \\
   & LSG$^{IJCAI'23}$~\cite{lian2023learnable} &  ResNet-19 & 4 & 95.17 \\
   & RateBP$^{NeurIPS'24}$~\cite{yu2024advancing} &  ResNet-18 & 4 & 95.61 \\
   & KDSNN$^{CVPR'23}$~\cite{KDSNN} &  ResNet-18 & 4 & 93.41 \\
   & Weak2Strong$^{NeurIPS'25}$~\cite{ding2025synergy} &  ResNet-19 & 4 & 96.66 \\
  \cline{2-5}
  \rowcolor{ourcolor} \cellcolor{white} &  & ResNet-18 & 4 & \textbf{95.70}\\
   \rowcolor{ourcolor} \cellcolor{white} & \multirow{-2}{*}{\textbf{Ours}} & ResNet-19 & 4 & \textbf{96.73}\\
  \hline
  \multirow{13}{*}{CIFAR100}
  & QKFormer$^{NeurIPS'24}$~\citep{QKFormer} & QKFormer & 4 & 81.15 \\ 
  & SNN-ViT$^{ICLR'25}$~\citep{wang2025spiking} & SNN-ViT & 4 & 80.10\\
  & RateBP$^{NeurIPS'24}$~\cite{yu2024advancing} &  ResNet-18 & 4 & 78.26 \\
  & STAA-SNN$^{CVPR'25}$~\citep{zhang2025staa} & ResNet-19 & 4 & 82.05 \\
  & DeepTAGE$^{ICLR'25}$~\citep{liu2025deeptage} & ResNet-19 & 4 & 81.39\\
  & RateBP$^{NeurIPS'24}$~\cite{yu2024advancing} &  ResNet-19 & 4 & 80.71 \\
   & TS-SNN$^{ICML'25}$~\citep{TSSNN} &ResNet-19 & 2 & 80.28 \\
   & ReverB-SNN$^{ICML'25}$~\citep{guo2025reverbsnn} & ResNet-19 & 2 & 78.46 \\
   & TMC$^{ICML'25}$~\citep{TMC} & ResNet-19 & 4 & 77.52 \\
   & SLT$^{AAAI'24}$~\citep{SLT} &  ResNet-19 & 4 & 75.01 \\
    & Weak2Strong$^{NeurIPS'25}$~\cite{ding2025synergy} &  ResNet-19 & 4 & 82.02 \\
  \cline{2-5}
   \rowcolor{ourcolor} \cellcolor{white} & & ResNet-18 & 4 & \textbf{78.83}\\
   \rowcolor{ourcolor} \cellcolor{white} &  \multirow{-2}{*}{\textbf{Ours}} & ResNet-19 & 4 & \textbf{82.29}\\
  \bottomrule
 \end{tabular}
 \end{threeparttable}
\end{table*}

\subsection{Further Comparison with MPS}

Due to MPS's strong performance on neuromorphic datasets, Table~\ref{com_MPS} provides an additional comparison between our method and MPS~\cite{MPS}. This comparison ensures that all training strategies, including model architecture and timestep settings, are identical. Our method demonstrated superior performance across three datasets, consistently outperforming MPS.

\subsection{Further Extension of the Proposed Method}

By default, we only apply the AND operation to the output of the penultimate layer (before the fully connected layer) to extract the stable pulse skeleton and compute the consistency loss. Naturally, we can extend this implementation to more layers. For now, we'll call it Stable Spike Dense. To accomplish this, we treat every two convolutions in VGG-9 as a stage and compute the consistency loss by performing an AND operation on the output features of each stage. As more layers have been optimized for consistency, the results in Table~\ref{com_dense} indicate further performance improvements.

\begin{table*}[t]
 \centering
 \tabcolsep=0.006\columnwidth
 \caption{Hyperparameter sensitivity experiments (\%) across architectures and timesteps on multiple datasets.}
 \label{com_sensitivity}
 \vskip -0.1in
 \begin{threeparttable}
 \scalebox{0.95}{
 \begin{tabular}{ccccc|cccccc|cccccccc}
  \toprule
 T=4,VGG-9,$\alpha=$  & 1 & 2 & 3  & 4 & $\beta=0.25$ & 0.5 & 0.75 & 1.0 & 1.25 & 1.5 & $\gamma= 0.25$ & 0.5 & 0.75 & 1.0 & 1.25 & 1.5\\
  \midrule
  CIFAR10-DVS  & 76.2  &  77.1 & 76.9 & 77.1 & 76.0 & 76.6 & \textbf{77.3} & 77.1 & 76.5 & 76.8 & 76.1 & 75.8 & 76.2 & 77.1 & 75.6 & 76.4\\
  N-Caltech101 &  82.39  & \textbf{83.92} & 82.49 & 82.82 & 82.71 & 82.60 & 82.60 & \textbf{83.92} & 83.15 & 82.82 & 83.15 & 83.26 & 82.71 & \textbf{83.92} & 82.71 & 82.28\\
  DVS-Gesture &  93.40 & 94.44 & \textbf{95.14} & 94.10 & 93.75 & 95.04 & 94.10 & 94.44 & 93.75 & 93.40  & 94.10 & 94.10 & 93.06 & 94.44 & 93.75 & 93.06\\
   \toprule
 T=16,QKFormer,$\alpha=$  & 1 & 2 & 3  & 4 & $\beta=0.25$ & 0.5 & 0.75 & 1.0 & 1.25 & 1.5 & $\gamma= 0.25$ & 0.5 & 0.75 & 1.0 & 1.25 & 1.5\\
  \midrule
  DVS-Gesture & 98.26 & \textbf{98.61} & 98.26 & 98.26 & 98.26 & 98.26 & 98.61 & 98.61 & 98.26 & 98.26 & 98.26 & 98.26 & 98.26 & 98.61 & 98.26 & 98.61\\
  \bottomrule
 \end{tabular}
 }
 \end{threeparttable}
 \vskip -0.1in
\end{table*}

\subsection{Experimental Results on CIFAR}
Table~\ref{com_cifar} shows the experimental results of our method using the ResNet-18 and ResNet-19 architectures on CIFAR10 and CIFAR100. With four timesteps, we achieved accuracy rates of 96.73\% on CIFAR10 and 82.29\% on CIFAR100, significantly outperforming other methods.

\subsection{Additional Hyperparameter Sensitivity Experiments}

To further investigate the sensitivity of the proposed method to hyperparameters, we conducted additional experiments across architectures and timesteps on neuromorphic datasets. Table~\ref{com_sensitivity} shows that our method exhibits stable overall performance as long as the hyperparameters remain within reasonable ranges ($\beta=\gamma=1.0$ and $\alpha=2.0$). We recommend increasing $\beta$ and $\gamma$ appropriately when tasks exhibit significant temporal fluctuations and increasing $\alpha$ appropriately when interclass differences are excessive to smooth out variations in probability distributions.

\subsection{Stable spikes can maintain temporal information.}
\textbf{Stable spikes maintain $T-1$ step temporal coherence by eliminating transient noise rather than erasing all temporal information.} Legitimate temporal dynamics are essentially the ordered evolution of spatial features over time. This evolution rarely isolates the features from adjacent timesteps, thus enabling their preservation by stable spikes. Additionally, Table~\ref{table:rebuttal_timestep} shows that optimizing only the first and last two timesteps results in suboptimal performance. This further validates that applying our method to all timesteps does not eliminate useful temporal information, but rather enhances it.

\begin{table}[!t]
 \centering
 \caption{Optimizing consistency across all timesteps does not eliminate useful temporal information.}
 \label{table:rebuttal_timestep}
 \tabcolsep=0.05\columnwidth
 \vskip -0.15in
 \begin{threeparttable}
 \scalebox{0.8}{
 \begin{tabular}{ccccc|}
  \toprule
 Consistency  & CIFAR10-DVS & DVS-Gesture \\
  \midrule
  \textbf{All timestep}  & \textbf{77.1} & \textbf{94.44}\\
  First two timesteps & 75.2 & 91.67\\
  Last two timesteps & 75.9 & 93.06\\
  \bottomrule
 \end{tabular}
 }
 \end{threeparttable}
 \vskip -0.18in
\end{table}

\begin{figure*}[t]
  \centering
  \includegraphics[width=0.8\linewidth]{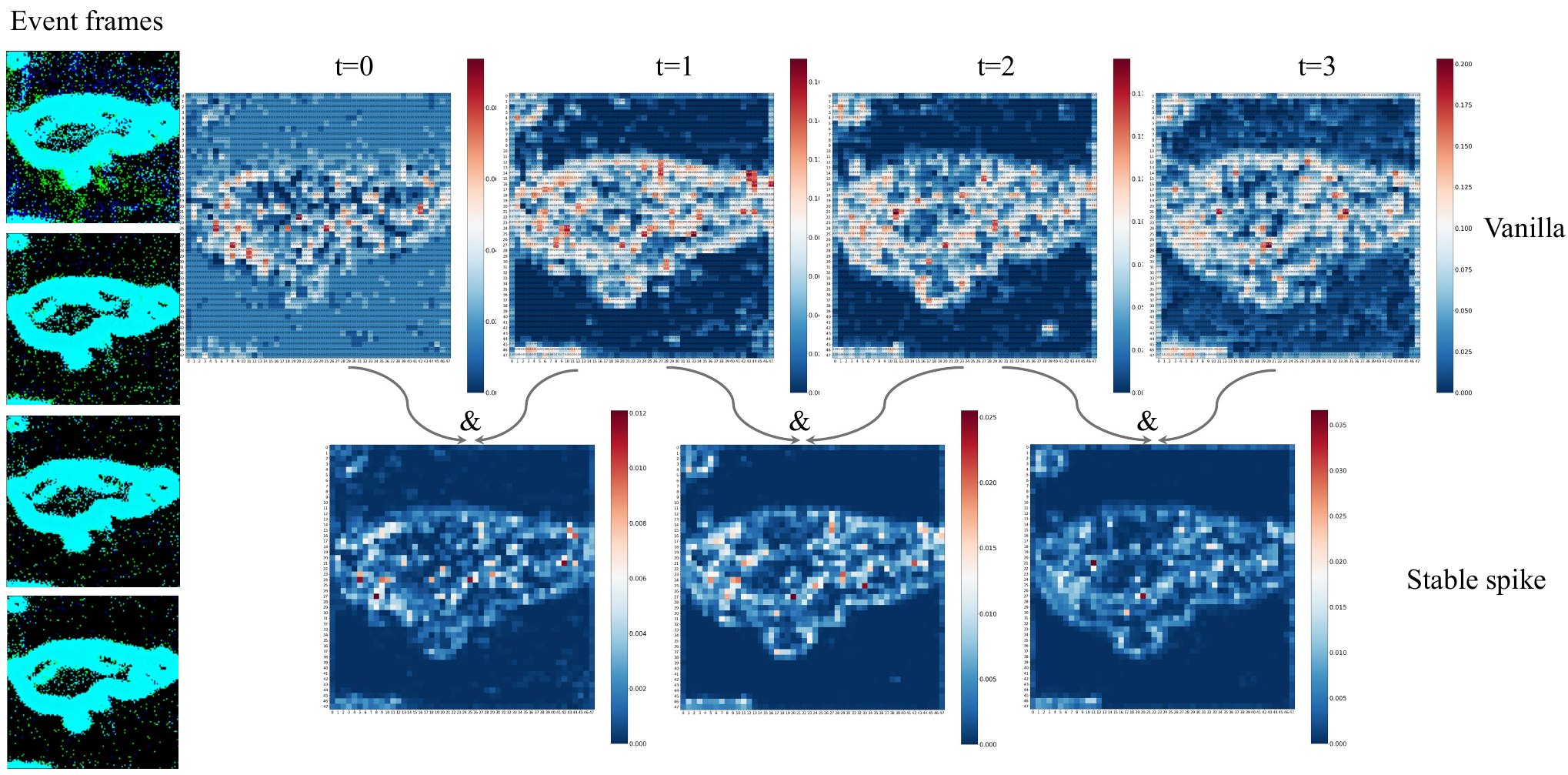}
  \caption{Visualization of the second layer spike maps of VGG-9 on CIFAR10-DVS. The stable spike maps decoupled by the minimal \& operation precisely depict the feature skeleton, echoing the results of Fig. 1 in the main paper. To ensure the visualization effect, the average of the spike maps of all channels is displayed.}
  \label{layer2cifar10dvs}
\end{figure*}

\begin{figure*}[h]
  \centering
  \includegraphics[width=0.8\linewidth]{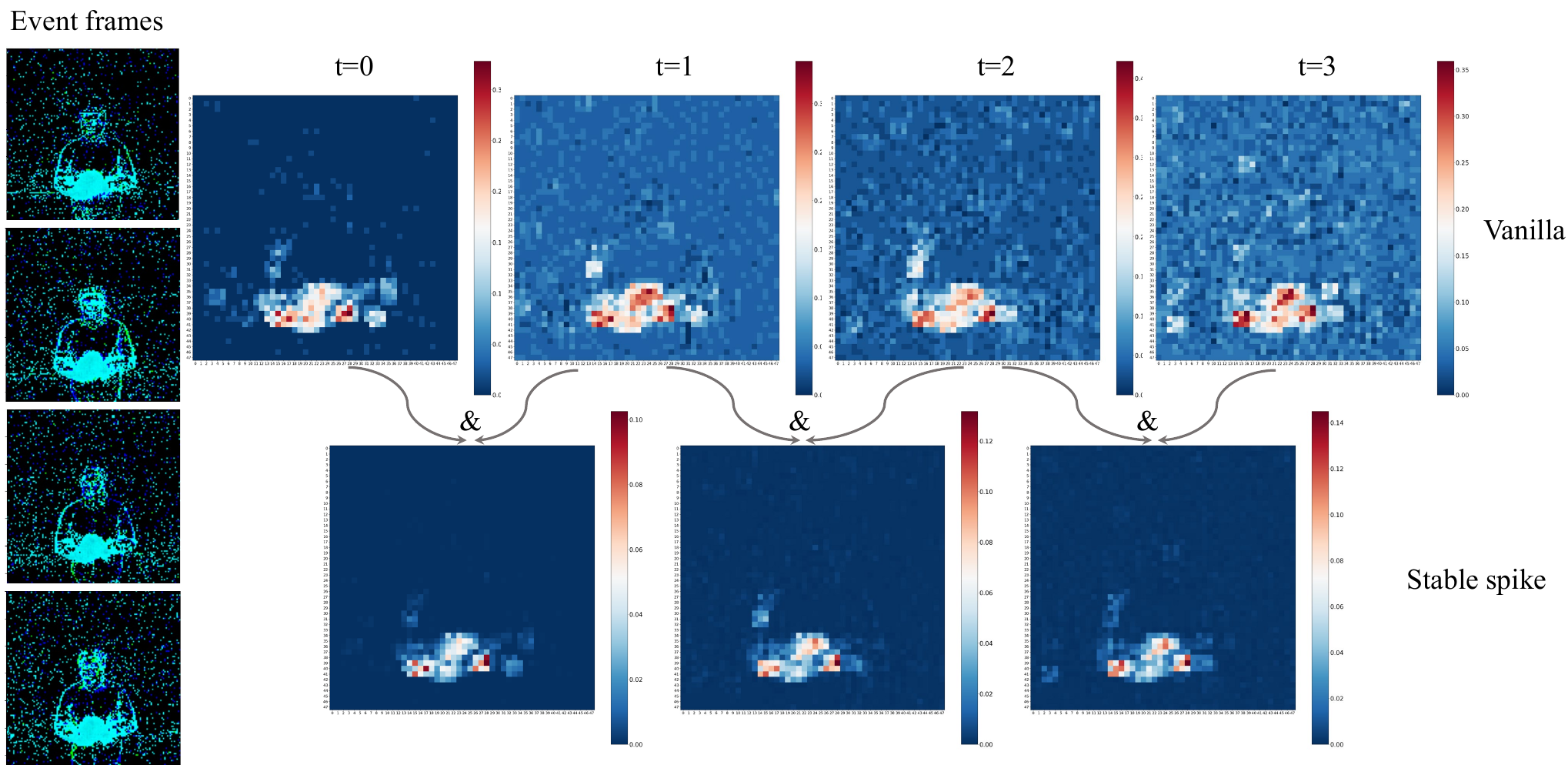}
  \caption{Visualization of the first layer spike maps of VGG-9 on DVS-Gesture. The stable spike maps decoupled by the minimal \& operation precisely depict the feature skeleton, echoing the results of Fig.1 in the main paper. To ensure the visualization effect, the average of the spike maps of all channels is displayed.}
  \label{layer1gesture}
\end{figure*}

\begin{figure*}[h]
  \centering
  \includegraphics[width=0.8\linewidth]{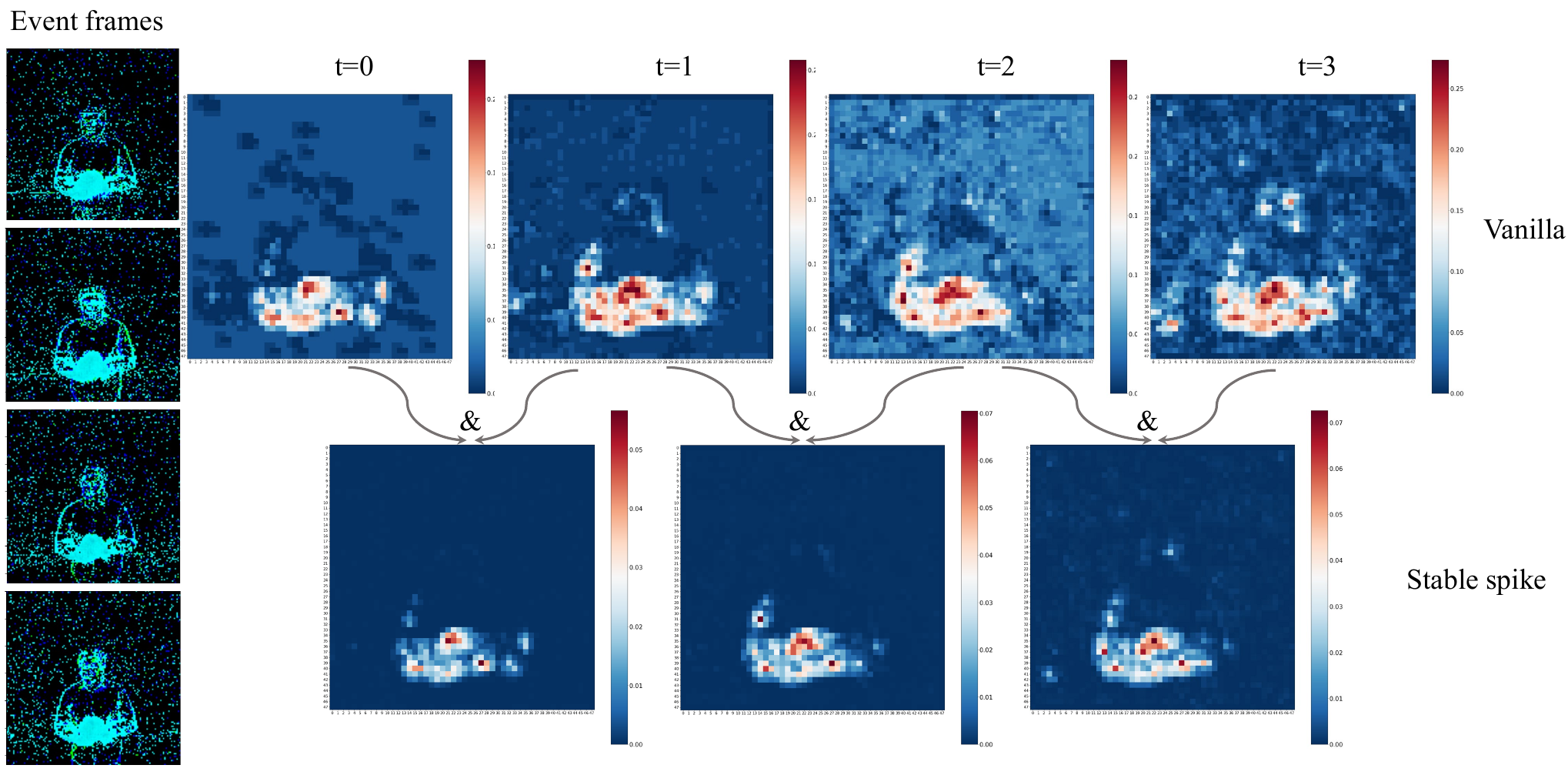}
  \caption{Visualization of the second layer spike maps of VGG-9 on DVS-Gesture. The stable spike maps decoupled by the minimal \& operation precisely depict the feature skeleton, echoing the results of Fig.1 in the main paper. To ensure the visualization effect, the average of the spike maps of all channels is displayed.}
  \label{layer2gesture}
\end{figure*}

\section{Negligible Training Overhead}
\label{overhead}

Our method only requires performing the AND operation on the last layer spike maps during training. Then, forward propagation is performed with noise added. The AND bit operation and the cost of generating random noise are negligible, and forward propagation only involves one fully connected layer. Therefore, we did not observe any significant memory or time overhead during training. During inference, our method is the same as the vanilla SNN, so it does not affect inference efficiency at all. Overall, our method significantly improves the inference performance of SNNs with negligible training overhead.

\newpage
\section{Additional Visualizations}
\label{add_vis}

In this section, we present additional spike map visualizations to demonstrate the consistency of ``the vanilla spike map differences across timesteps and the ability of the stable spike to accurately represent the feature skeleton" across multiple datasets and layers. Fig. 1 in the main paper shows the spike map of the first layer of the VGG-9 architecture on CIFAR10-DVS. Fig.~\ref{layer2cifar10dvs} shows the spike map of the second layer. Fig.~\ref{layer1gesture} and Fig.~\ref{layer2gesture} show the spike maps of the first two layers on DVS-Gesture, respectively.

\end{document}